Research paper

# Material-agnostic temperature field prediction for metal additive manufacturing via a parametric PINN framework

Hyeonsu Lee, Jihoon Jeong *

*Wm Michael Barnes '64 Department of Industrial and Systems Engineering, Texas A&M University, College Station, TX 77843, USA*

ABSTRACT

Accurate temperature field prediction in metal additive manufacturing (AM) is essential for understanding the process–structure–performance relationship. While prior studies have explored generalization to unseen process conditions, they often require extensive datasets, costly retraining, or pre-training. Generalization across different materials also remains relatively unexplored due to the challenges posed by distinct material-dependent thermal behaviors. This paper introduces a parametric physics-informed neural network (PINN) framework for generalization across unseen materials without labeled data, retraining, or pre-training. The framework adopts a decoupled parametric PINN architecture that separately encodes material properties and spatiotemporal coordinates, fusing them through conditional modulation to better align with the multiplicative role of material parameters in the governing equation and boundary conditions. Physics-guided output scaling derived from Rosenthal's analytical solution and a hybrid optimization strategy are further incorporated to enhance physical consistency, training stability, and convergence. Experiments with numerical simulations across diverse metal alloys, including both in-distribution and out-of-distribution cases, demonstrate effective generalizability along with superior training efficiency. Specifically, the proposed framework achieved up to a 64.2% reduction in relative $L_2$ error compared to the non-parametric baseline while surpassing its performance within only 4.4% of the baseline training epochs. Ablation studies clarify each component's contribution and scrutinize the severe training instability prevalent in conventional parametric PINNs. Overall, the proposed framework provides an efficient and scalable material-agnostic solution for temperature field modeling, contributing to more flexible and practical deployment in metal AM.

## 1. Introduction

In recent years, metal additive manufacturing (AM) has gained significant attention due to its potential across diverse applications ranging from aerospace, automotive, and agricultural components to construction and infrastructure [1]. The primary advantage of metal AM lies in its flexibility to realize not only complex geometric variations, but also material compositions tailored to the unique characteristics of each application [2]. Notwithstanding this considerable potential, the prevalence of process-induced defects and the difficulty in ensuring consistent material performance remain critical bottlenecks to widespread industrial adoption [3].

To overcome these obstacles, understanding thermal history plays a critical role. Accurate thermal field information is indispensable for understanding defect formation and the process-structure–property chain [4,5]. To this end, traditional analytical methods [6,7] and numerical methods such as the Finite Element Method (FEM) [8,9] have been employed for thermal modeling in metal AM. However, analytical methods often struggle to account for the multi-scale physics and complex boundary conditions inherent in AM, while numerical approaches are often constrained by high computational cost and limited flexibility for rapid design iterations [10,11].

*Data-driven methods for thermal modeling.* Data-driven methods using machine learning (ML) have been proposed as surrogate models to complement these numerical methods [12], and have succeeded in rapid temperature prediction for diverse scenarios [13,14]. Such methods have received significant attention due to their rapid prediction capability and modeling flexibility, with recent advances focusing not only on improving prediction accuracy [15–17], but also on *enhancing generalization to unseen process conditions* [18,19]. Specifically, various data-driven methods have been explored to generalize across unseen geometries [20], tool paths [21,22], and process parameters [23]. Nevertheless, generalization to material variations remains underexplored. To the best of the authors' knowledge, this has been explored only by a transfer-learning approach that pre-trains on extensive source-domain data and fine-tunes on target substrate materials [24].

* Corresponding author.
*E-mail addresses:* hslee@tamu.edu (H. Lee), jihoonjeong@tamu.edu (J. Jeong).

Despite these notable advances, purely data-driven models face several critical shortcomings. As black-box models, they often lack physical interpretability and reliability, which may hinder their applicability in practice [25]. Furthermore, they rely heavily on large volumes of high-fidelity ground-truth data for supervision; since experimental data are inherently limited, training these models in practice still relies on the very numerical simulations they originally aimed to replace [26]. Two further limitations constrain their generalization: such approaches require extensive data generation for both pre-training and fine-tuning, incurring excessive computational overhead, and adaptation through fine-tuning requires prior exposure to the target scenario before inference, thereby limiting generalization capability.

*PINN-based thermal field prediction.* Physics-Informed Neural Networks (PINNs) have emerged as a promising solution to address the aforementioned limitations by embedding governing physical laws, such as partial differential equations (PDEs), initial conditions, and boundary conditions, directly into the loss function of ML models [27]. Even though PINNs often encounter severe training instabilities and optimization complexities compared to purely data-driven models [28], they offer the distinct advantages of mesh-free architectures and data-efficient, or even *data-free* training, positioning PINNs as a compelling framework for thermal modeling with *generalization to unseen process conditions*.

Existing PINN-based thermal studies can be broadly categorized along two directions: (1) improving performance by mitigating optimization challenges, and (2) enhancing generalization to unseen process conditions. From an optimization perspective, PINNs are known to suffer from significant training instability due to multi-objective formulations and higher-order derivatives [28]; to attenuate this instability in metal AM, various approaches have been proposed, including adaptive loss weighting and non-dimensionalization [29], domain decomposition [30], ensemble-based error homogenization [31], and advanced feature encoding methods [32].

From a generalization perspective, most existing PINN frameworks have focused on fast adaptation rather than rapid inference. Because a substantial body of conventional PINNs remains largely non-parametric, they are typically optimized for a specific process condition with a fixed set of PDE parameters. Consequently, any change in these parameters for a new process condition necessitates retraining the model from scratch to re-satisfy the governing PDEs. To alleviate this computational burden, transfer-learning strategies have been introduced to accelerate adaptation to new process-parameter settings [26], and later works have extended this framework to accommodate variations in deposition patterns, laser paths, and beam profiles [33–35]. Nevertheless, as with the data-driven methods discussed above, these approaches still require per-condition adaptation and prior exposure to target conditions before inference, limiting the scalability of PINN frameworks and potentially undermining the flexibility that metal AM promises.

Parametric PINNs offer a fundamentally different route to generalization. Rather than treating the PDE parameters as fixed constants, they take these parameters as additional network inputs. A single training run then amortizes an entire family of parameterized PDEs, returning the solution for a new process condition in a single forward pass without retraining, fine-tuning, or any target-domain data. Building on this idea, a parametric formulation has recently been proposed that takes both spatiotemporal coordinates and ranges of process parameters and material properties as inputs [36]. Process parameters have likewise been incorporated as inputs, in combination with transfer learning [37] for rapid inference across the unseen process conditions with respect to the changes in process parameters.

Although these recent advances demonstrate that parametric PINNs are conceptually well-suited to generalization across unseen process conditions [36,37], several limitations remain open to further investigation:

(1) **Cross-material generalization remains partially addressed.** To the best of the authors' knowledge, prior literature addressing cross-material variations [36] has primarily trained and evaluated performance within a relatively narrow band of thermophysical properties (e.g., thermal conductivity in $[5, 20]\,\mathrm{W\,m^{-1}\,K^{-1}}$). However, diverse materials utilized in metal AM typically exhibit wide variations in thermophysical properties, leading to distinct thermal behaviors and substantial variations in peak temperatures [38]. Consequently, generalization to materials with significantly different characteristics, particularly in out-of-distribution (OOD) regimes beyond the training range, remains less understood. Furthermore, our findings suggest that such material-induced variations in thermal scale introduce noticeable training instabilities in parametric PINNs, presenting an optimization challenge that warrants further attention.

(2) **The advantage of parametric formulations over non-parametric PINNs is not yet well understood.** Because comparisons against non-parametric baselines are often omitted in prior works [36,37], it remains unclear whether parametric formulations provide actual gains in accuracy or scalability beyond generalization. Perhaps rapid inference across unseen conditions in parametric PINNs offers limited benefit if it comes at the cost of accuracy compared to non-parametric PINNs. Critically, our experiments show that previous parametric PINN approaches with the monolithic architecture, which concatenate spatiotemporal coordinates and PDE parameters into a single input vector, often fail to outperform non-parametric baselines. Moreover, we find that these monolithic parametric PINNs are often prone to severe training instability, which we attribute in part to a misalignment between physics and representation.

(3) **Extremely long training horizons remain a major practical challenge.** Due to the inherent optimization complexities and steep loss landscapes in PINNs, many approaches utilize small step sizes and extensive training horizons, requiring tens or hundreds of thousands of epochs to attenuate training instability and ensure stable convergence. However, this severe computational overhead may present a critical bottleneck for adopting PINNs in practice.

*Main contributions.* To address these limitations, this paper introduces a parametric PINN framework for material-agnostic temperature field prediction in metal AM. Compared to prior PINN works, our framework is designed to handle a wide range of materials used in metal AM, severe training instability induced by the large variations in temperature field magnitude across materials' thermophysical properties, and the extremely long training horizons typical in conventional PINNs. To this end, the proposed framework is built upon three key components:

- **Decoupled parametric architecture.** We develop a decoupled architecture for metal AM so that spatiotemporal coordinates and material properties are processed through separate branches before fusing them. Beyond merely leveraging the architecture of [39], we further clarify the source of its superiority over the monolithic architecture, especially in the cross-material generalization context. Specifically, by realizing a fusion mechanism analogous to feature-wise linear modulation (FiLM) [40], material properties modulate the spatiotemporal features, imposing a physics-consistent inductive bias that leads to better representation learning.
- **Physics-guided output scaling.** We discover that parametric PINNs for cross-material generalization often suffer from severe training instability, as the temperature magnitude varies substantially across materials with distinct thermophysical properties. To mitigate this, we introduce a physics-guided output scaling strategy grounded in Rosenthal's analytical solution. Inspired by the similarities between welding processes and metal AM, we effectively approximate the peak temperatures and scale distinct temperature field magnitudes across materials in a principled and scalable way, thereby stabilizing training for cross-material generalization.

- **Hybrid optimization strategy.** We combine first- and second-order optimization to better condition the stiff loss landscape of PINNs and adapt efficiency in the use of the second-order optimizer, accelerating convergence and overcoming the extensive training horizons prevalent in conventional PINNs in metal AM.

Extensive numerical experiments demonstrate the proposed framework's strong cross-material generalization in both in-distribution and challenging OOD regimes, with higher accuracy and training efficiency than previous PINN approaches — including both non-parametric baselines and conventional monolithic parametric PINNs — by preventing material-specific overfitting and better aligning with the underlying physics in metal AM. Beyond quantitative evaluations, we provide qualitative analyses that reveal how different thermophysical properties across materials shape parametric PINN-based thermal modeling. Furthermore, additional studies clarify each component's contribution to overall performance and scrutinize the sources of severe training instability in parametric PINNs. Finally, we show that the components of the proposed framework are broadly applicable to other PINN formulations in metal AM beyond the present framework.

Note that this work belongs to a pure physics-driven forward-modeling paradigm, predicting the temperature field across unseen material changes in metal AM without any ground-truth data, retraining, or fine-tuning. We further note the remarkable progress of recent state-of-the-art frameworks that tackle far more challenging, strongly coupled multi-physics problems [41] and employ advanced active-learning strategies to dramatically reduce data requirements during transfer learning [42]. These works represent particularly significant advances for PINN-based modeling, yet their application scenarios differ somewhat from ours.

## 2. Methodology

### 2.1. Nomenclature

We leave the nomenclature in Appendix A.1.

### 2.2. Problem description

We begin by defining material-agnostic temperature prediction as a generalization problem over material-dependent thermal fields governed by the same underlying physical laws. Let the spatiotemporal domain be denoted by $\mathcal{D} := \Omega \times [0, t_{\text{end}}]$, where $\Omega \subset \mathbb{R}^d$ is the spatial domain and $(\mathbf{x}, t) \in \mathcal{D}$. Each material is characterized by an $M$-dimensional vector of material properties (e.g., density, heat capacity, and thermal conductivity), denoted by $\lambda \in \mathcal{M} \subset \mathbb{R}^M$, where $\mathcal{M}$ represents the admissible material property space and each component $\lambda_m$ lies within known bounds. The corresponding temperature field is defined as

$$\mathbb{T} := \{T(\mathbf{x}, t, \lambda) \in \mathbb{R} \mid (\mathbf{x}, t) \in \mathcal{D},\ \lambda \in \mathcal{M}\}$$

Our goal is to learn a predictive model $\mathcal{F}_\Theta : \mathcal{D} \times \mathcal{M} \to \mathbb{T}$, parameterized by the set of all learnable parameters $\Theta$, that minimizes the expected generalization error across both the spatiotemporal domain and the material property space. Formally, the objective can be expressed as

$$\min_{\Theta} \mathbb{E}_{\lambda \sim p_{\mathcal{M}}} \left[ \mathbb{E}_{(\mathbf{x},t) \sim p_{\mathcal{D}}} \left[ \left\| T(\mathbf{x}, t, \lambda) - \hat{T}_\Theta(\mathbf{x}, t, \lambda) \right\|_2^2 \right] \right] \tag{1}$$

In practice, however, obtaining ground-truth temperature fields $T(\mathbf{x}, t, \lambda)$ over the entire $\mathcal{D} \times \mathcal{M}$ input domain is infeasible due to the extensive computational cost of performing numerical simulations for every material. As a result, purely data-driven supervision may not be effective. In this sense, we adopt PINNs, which incorporate governing physical laws as constraints during training. By embedding physics-based inductive bias directly into the learning objective, PINNs not only promote physically consistent and interpretable predictions but also enable training even in the absence of ground-truth temperature data.

Therefore, the objective in Eq. (1) can be reformulated as the minimization of physics-based residuals:

$$\min_{\Theta} \mathbb{E}_{\lambda \sim p_{\mathcal{M}}} \left[ \mathbb{E}_{(\mathbf{x},t) \sim p_{\mathcal{D}}} \left[ \mathcal{L}_\Theta(\mathbf{x}, t, \lambda) \right] \right] \tag{2}$$

subject to $\mathcal{L}_\Theta(\mathbf{x}, t, \lambda) = w_{\text{pde}} \mathcal{L}_{\text{pde}} + w_{\text{bc}} \mathcal{L}_{\text{bc}} + w_{\text{ic}} \mathcal{L}_{\text{ic}}$

To satisfy the governing PDEs, boundary conditions, and initial conditions, each loss is defined respectively as:

$$\mathcal{L}_{\text{pde}} = \frac{1}{N_{\text{pde}}} \sum_{n=1}^{N_{\text{pde}}} \left\| \mathcal{N}\left(\hat{T}_{\text{phys}}(\mathbf{x}^{(n)}, t^{(n)}, \lambda^{(n)}; \Theta)\right) \right\|_2^2, \quad (\mathbf{x}^{(n)}, t^{(n)}) \in \Omega \times (0, t_{\text{end}}],\ \lambda^{(n)} \in \mathcal{M} \tag{3}$$

$$\mathcal{L}_{\text{bc}} = \frac{1}{N_{\text{bc}}} \sum_{n=1}^{N_{\text{bc}}} \left\| \mathcal{B}\left(\hat{T}_{\text{phys}}(\mathbf{x}^{(n)}, t^{(n)}, \lambda^{(n)}; \Theta)\right) \right\|_2^2, \quad (\mathbf{x}^{(n)}, t^{(n)}) \in \partial\Omega \times (0, t_{\text{end}}],\ \lambda^{(n)} \in \mathcal{M} \tag{4}$$

$$\mathcal{L}_{\text{ic}} = \frac{1}{N_{\text{ic}}} \sum_{n=1}^{N_{\text{ic}}} \left\| \mathcal{I}\left(\hat{T}_{\text{phys}}(\mathbf{x}^{(n)}, 0, \lambda^{(n)}; \Theta)\right) \right\|_2^2, \quad \mathbf{x}^{(n)} \in \Omega,\ \lambda^{(n)} \in \mathcal{M} \tag{5}$$

Here, $\mathcal{N}$, $\mathcal{B}$, and $\mathcal{I}$ are the residual operators of the PDE, boundary, and initial conditions, and $N_{\text{pde}}$, $N_{\text{bc}}$, and $N_{\text{ic}}$ are the numbers of collocation points for each. In this study, the governing PDE is the full 3D transient heat diffusion equation with Fourier conduction, coupled with convective, radiative, and Gaussian laser-flux boundary conditions, and an ambient temperature initial condition. More detailed mathematical formulations are deferred to Appendix A.3.

It is notable that there is no data-driven loss term, as preparing ground-truth temperature data for all materials is infeasible. This, in turn, makes our problem more challenging compared to hybrid settings, where auxiliary data-consistency losses (e.g., Eq. (1)) often guide the minimization of physics-based residuals [26]. Consequently, the optimization landscape is more prone to instability and convergence to suboptimal solutions.

In addition, we assume fixed process parameters, fixed part geometry, and a material-independent collocation sampling strategy throughout this study. Under this assumption, the process parameters as well as the spatiotemporal collocation domain and its distribution remain invariant across different materials, and only the material property vector $\lambda$ varies. While several studies have explored process parameter-aware temperature prediction models [33,37], we leave relaxing these assumptions to future work, as the aim and scope of this paper are to build an effective temperature prediction model for unseen materials.

### 2.3. Methodology overview

Fig. 1 illustrates the overall proposed framework. Specifically, we first predict the raw temperature field using a parametric PINN with a decoupled architecture. We then incorporate physics-guided output scaling derived from Rosenthal's analytical solution, along with a hybrid training strategy, to enhance physical consistency, training stability, and convergence of parametric PINNs for material-agnostic temperature prediction.

### 2.4. Revisiting the architecture in parametric PINNs: Decoupled architecture

Existing parametric PINNs in metal AM typically adopt their model $\mathcal{F}_\Theta$ with a monolithic architecture that treats spatiotemporal points and parametric inputs as a single input vector [36,37]. However, we suggest that such formulations may not adequately capture the relationship between the spatiotemporal point $(\mathbf{x}, t)$ and the material property vector $\lambda$ when predicting the temperature field $T(\mathbf{x}, t, \lambda)$, especially in the cross-material thermal modeling scenario.

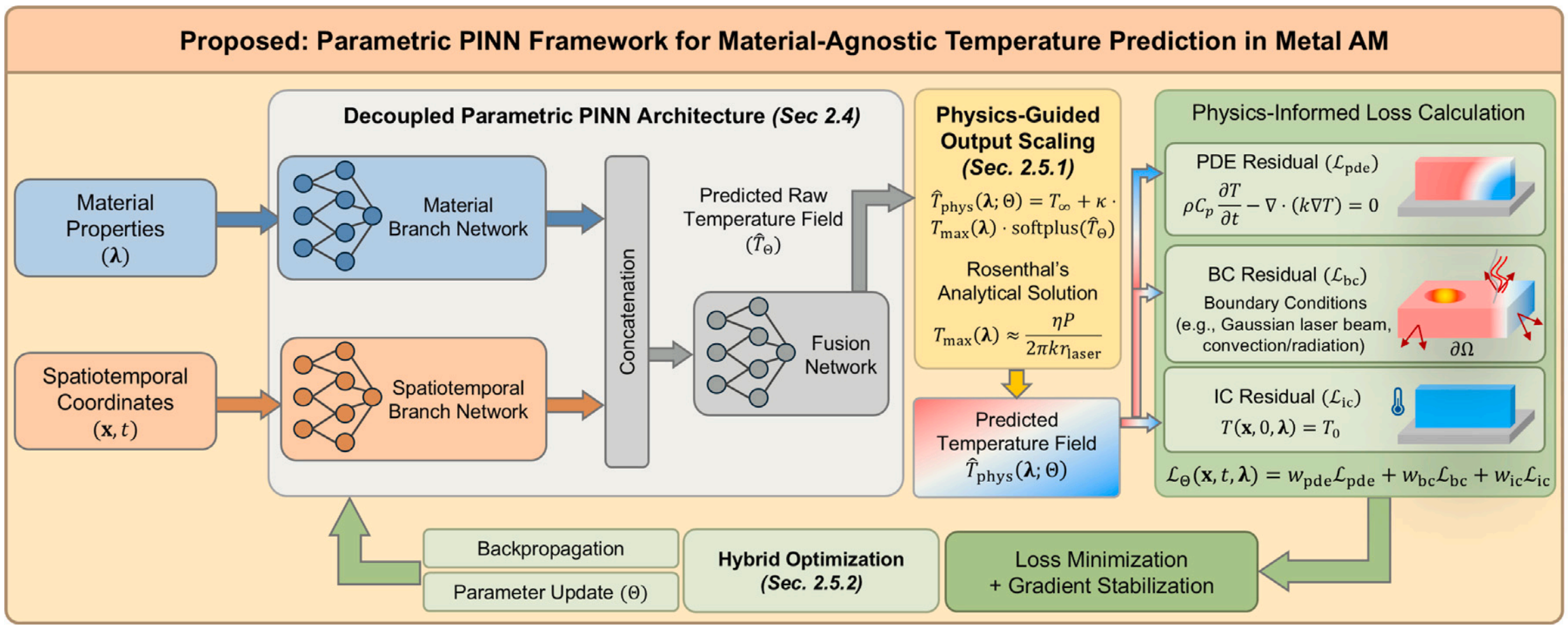

**Fig. 1.** Proposed parametric PINN framework for material-agnostic temperature prediction in metal AM.

Let us denote such a monolithic prediction network as $g_\Theta : (\mathbf{x}, t, \lambda) \mapsto \hat{T}_\Theta$. Typically, $g_\Theta$ is implemented as a fully connected neural network consisting of $L$ hidden layers $\{\mathbf{h}_l\}_{l=1}^{L}$, which are defined recursively as follows:

$$g_\Theta(\mathbf{x}, t, \lambda) = \mathrm{h}_L(\mathrm{h}_{L-1}(\cdots(\mathrm{h}_1(\mathbf{x}, t, \lambda))))$$

Here, the operation within the first hidden layer $\mathrm{h}_1$ is expressed as:

$$\mathrm{h}_1 = \psi(\mathrm{W}_{1,\mathbf{x}}\mathbf{x} + \mathrm{W}_{1,t}t + \mathrm{W}_{1,\lambda}\lambda + \mathrm{b}_1) \tag{6}$$

where $\psi(\cdot)$ is an activation function, $\mathrm{W}_{1,\mathbf{x}}, \mathrm{W}_{1,t}, \mathrm{W}_{1,\lambda}$ represent the weight matrices for the spatial, temporal, and parameter inputs respectively, and $\mathrm{b}_1$ is the bias term.

While this formulation is straightforward, it may be suboptimal for capturing the underlying physics of the problem from two perspectives. From a physical perspective, under the assumption of temperature-independent material properties adopted in this study, material properties typically do not vary across the spatiotemporal domain; rather, they act as *conditional parameters* that govern thermal behavior over the entire domain. From a training perspective, material parameters in most governing PDEs and boundary conditions in metal AM appear as multiplicative coefficients of field derivatives.

However, as shown in Eq. (6), a monolithic architecture initializes interactions between the spatiotemporal inputs $(\mathbf{x}, t)$ and the material properties $\lambda$ through a linear additive combination. Consequently, the network must implicitly approximate these multiplicative physical relationships using summation-based representations, leading to inefficient representation learning. This, in turn, forces the network to approximate inherently multiplicative interactions through deeper nonlinear compositions, increasing sample complexity and hindering generalization across materials. This observation motivates the design of a more effective parametric PINN architecture tailored for material-agnostic temperature prediction.

To overcome this limitation, we develop a decoupled architecture that processes spatiotemporal features and material parameters separately prior to fusion. The architecture consists of two independent feature extraction networks followed by a fusion network. Fig. 2 illustrates the structural difference between the conventional parametric PINN $g_\Theta$ and the proposed decoupled architecture. The network is defined as:

$$\mathcal{F}_\Theta(\mathbf{x}, t, \lambda) = g_{\theta_c}\left[g_{\theta_{\mathbf{x},t}}(\mathbf{x}, t), g_{\theta_\lambda}(\lambda)\right] \tag{7}$$

where $g_{\theta_{\mathbf{x},t}}$ and $g_{\theta_\lambda}$ denote the sub-networks for spatiotemporal coordinates and material properties, respectively, and $g_{\theta_c}$ represents the fusion network. The complete set of learnable parameters is given by $\Theta = \{\theta_c, \theta_{\mathbf{x},t}, \theta_\lambda\}$.

This decoupled design enables a more structured modeling of the interaction between spatiotemporal coordinates and material properties. While the original study [39] provided limited justification for this parametric PINN architecture, we present a deeper rationale for its effectiveness compared to a monolithic design, particularly in cross-material thermal modeling scenarios.

Specifically, the fusion mechanism in the proposed architecture operates analogously to Feature-wise Linear Modulation (FiLM) [40]. Under this interpretation, the material property branch $g_{\theta_\lambda}$ generates conditional modulation parameters rather than merely serving as an additional input. Accordingly, the operation in Eq. (7) can be interpreted as approximating

$$\hat{T}_\Theta(\mathbf{x}, t, \lambda) \approx \gamma(\lambda) \odot g_{\theta_{\mathbf{x},t}}(\mathbf{x}, t) \oplus \zeta(\lambda)$$

where $\gamma(\lambda)$ and $\zeta(\lambda)$ denote learned scaling and shifting functions, respectively. In this formulation, $\lambda$ acts as a conditional variable that implicitly modulates spatiotemporal features in a multiplicative manner. This mechanism is more consistent with the role of material parameters in governing PDEs and boundary conditions, where they typically appear as multiplicative coefficients of field derivatives.

In this way, $\lambda$ functions as a conditional variable that dynamically adjusts the temperature profile at a given spatiotemporal point $(\mathbf{x}, t)$. This multiplicative interaction aligns more closely with the physical nature of the problem, enabling improved generalization across materials. As demonstrated in the experimental results, the proposed decoupled architecture consistently outperforms conventional parametric PINNs, despite employing fewer trainable parameters and lower overall network complexity.

## 2.5. Overcoming training instability and accelerating convergence

We present our two strategies designed for parametric PINNs in the material-agnostic temperature prediction scenario.

### 2.5.1. Physics-guided output scaling

In practical PINN formulations in metal AM, the raw network output $\hat{T}_\Theta$ is not often directly enforced to satisfy the governing PDE, boundary and initial conditions. Instead, it is typically transformed into a physically admissible temperature field $\hat{T}_{\text{phys}}(\Theta)$ through an output scaling mechanism [26,34]. A commonly adopted construction in prior studies is given by

$$\hat{T}_{\text{phys}}(\mathbf{x}, t; \Theta) \triangleq T_\infty + T_{\max} \cdot \underbrace{\mathrm{Softplus}(\hat{T}_\Theta)}_{\in[0,\infty)} \tag{8}$$

where $\hat{T}_\Theta = \mathcal{F}_\Theta(\mathbf{x}_{\text{norm}}, t_{\text{norm}})$

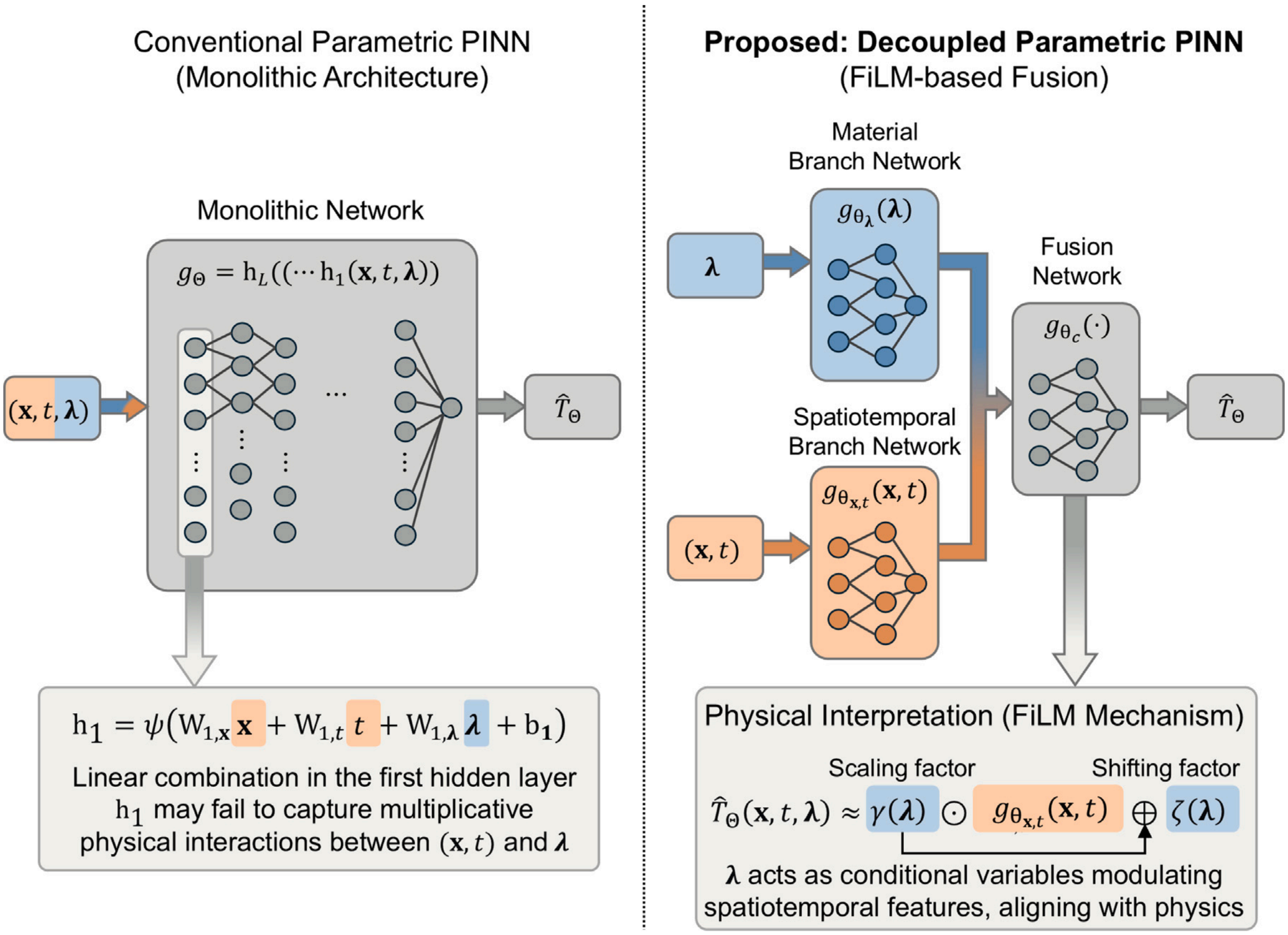

**Fig. 2.** Architectural comparison between a conventional monolithic parametric PINN and the proposed decoupled parametric PINN. The monolithic architecture jointly processes spatiotemporal coordinates and material properties through early fusion, which may limit its ability to represent physical interactions between spatiotemporal variables and material properties. In contrast, grounded in the FiLM mechanism [40], the proposed framework decouples material and spatiotemporal feature learning and fuses them through conditional modulation, allowing material properties to scale and shift spatiotemporal features in a manner consistent with physical interpretation.

Here, $T_\infty$ denotes the ambient temperature, and the activation function $\mathrm{Softplus}(u) = \log(1+e^u)$ enforces non-negativity while remaining smooth and differentiable. $(\mathbf{x}_{\mathrm{norm}}, t_{\mathrm{norm}})$ are normalized spatiotemporal inputs to stabilize gradient-based optimization, which is particularly important in PINNs [43].

This output scaling serves two fundamental purposes. First, under the standard assumption in metal AM that the initial temperature equals the ambient temperature [30,33] (i.e., $T_0 = T_\infty$), the lower bound of the Softplus function guarantees that the final predicted temperature $\hat{T}_{\mathrm{phys}}(\Theta)$ does not fall below the ambient temperature $T_\infty$ and initial temperature $T_0$ anywhere in the spatiotemporal domain, thereby enforcing the physically admissible temperature range. Second, and more importantly from an optimization perspective, the scaling factor $T_{\max}$ directly constrains the magnitude of the predicted raw temperature field $\hat{T}_\Theta$ and consequently controls the scale of gradients propagated through the governing PDE residuals, as well as the boundary and initial condition residuals.

From a training standpoint, introducing $T_{\max}$ can be interpreted as an explicit mechanism for gradient scale regulation. PINNs are known to be particularly sensitive to gradient imbalance due to the simultaneous optimization of multiple physics-based loss terms involving higher-order derivatives [28]. Without such output scaling in thermal modeling for metal AM, besides ensuring physical consistency, large disparities in gradient magnitudes can arise, leading to gradient explosion or vanishing and, consequently, severely hindering convergence. By narrowing the effective output range of the network prior to enforcing physical constraints on the loss function, Eq. (8) alleviates this burden and stabilizes the optimization landscape.

Empirical evidence from prior literature [44] indicates that removing this scaling factor $T_{\max}$ altogether, i.e.,:

$$\hat{T}_{\mathrm{phys}}(\mathbf{x}, t; \Theta) = T_\infty + \mathrm{Softplus}(\hat{T}_\Theta) \tag{9}$$

leads to significant training instability and poor convergence. This suggests that the primary role of $T_{\max}$ lies not merely in ensuring physical consistency with the governing equations, but more critically in stabilizing the gradient magnitudes encountered during optimization.

However, stabilizing the gradient scale becomes more fundamentally challenging in material-agnostic temperature prediction, where $T_{\max}$ must be specified for any unseen material properties $\lambda$. Prior studies have typically determined $T_{\max}$ through manual trial-and-error, which is neither scalable nor principled. Despite its critical role in stable PINN training, particularly in cross-material generalization settings, the selection of $T_{\max}$ remains largely underexplored in the literature. This motivates the need for a physics-informed, material-dependent approximation of the temperature upper bound.

To address this challenge, we employ Rosenthal's analytical solution to approximate the upper bound of the material temperature. In metal AM, the peak temperature $T_{\max}$ can be estimated using this solution; while originally developed for fusion welding, it is often adopted in metal AM due to the fundamental process similarities [45,46].

According to Rosenthal's analytical solution, the thermal history at a given point is expressed as:

$$T = T_0 + \frac{\eta P}{2\pi k r_{\mathrm{laser}}} \exp\left(-\frac{v(r_{\mathrm{laser}} + \xi)}{2\alpha}\right) \tag{10}$$

where the process parameters $\eta, P, r_{\mathrm{laser}}, v$, and $\xi (= x - vt)$ denote the laser absorptivity, power, beam radius, scanning speed, and relative position, respectively. The material properties are defined by thermal conductivity $k$ and thermal diffusivity $\alpha(=\frac{k}{\rho C_p})$. Since the exponential term is bounded by unity, the theoretical maximum temperature rise is simplified as:

$$T_{\max} = T_0 + \frac{\eta P}{2\pi k r_{\text{laser}}} \tag{11}$$

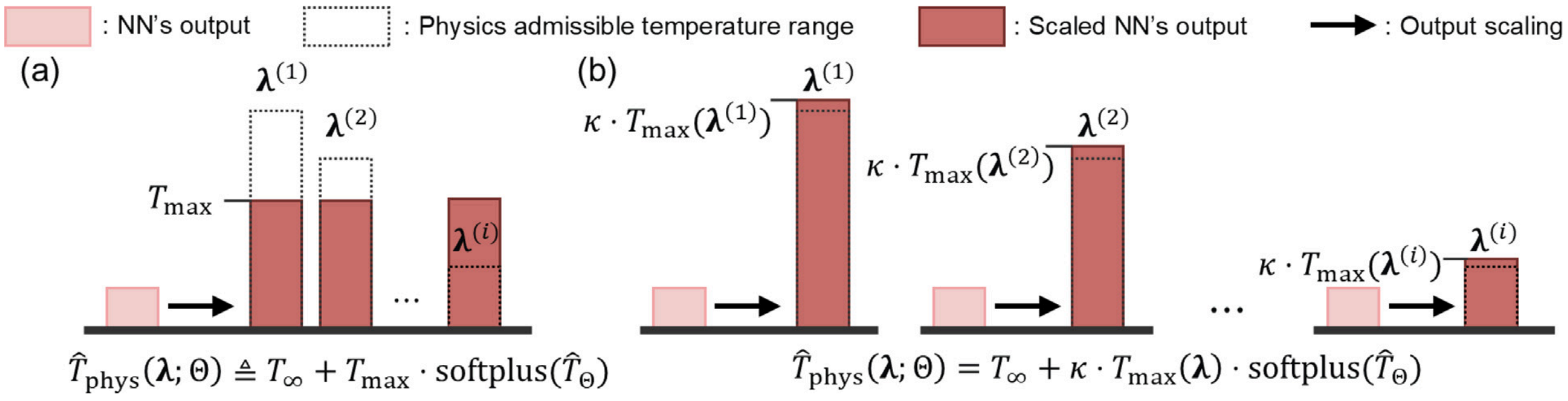

**Fig. 3.** Comparison of output scaling strategies for temperature prediction. (a) Manual output scaling [26] uses a manually determined, fixed peak temperature $T_{\max}$, failing to generalize across unseen materials $\lambda$. (b) Our physics-guided output scaling derives $T_{\max}(\lambda)$ from Rosenthal's analytical solution, enabling generalization to unseen material while ensuring physical consistency and stabilizing the optimization landscape by regulating gradient scales.

Although this approximation is physics-based, the actual maximum temperature may vary in metal AM. The analytical solution assumes a quasi-steady state and a point/line heat source, whereas actual metal AM processes involve highly transient thermal gradients and complex melt-pool dynamics that often lead to higher peak temperatures. To compensate for the inherent underestimation of Rosenthal's model when applied to metal AM scenarios, we introduce a correction factor $\kappa \geq 1$ and improve Eq. (8) by defining the *physics-guided output scaling* law as follows:

$$\begin{aligned} &\hat{T}_{\text{phys}}(\mathbf{x}, t, \lambda; \Theta) \triangleq T_\infty + \kappa \cdot T_{\max}(\lambda) \cdot \text{Softplus}\left(\hat{T}_\Theta\right) \\ &\quad \text{where} \quad \hat{T}_\Theta = \mathcal{F}_\Theta(\mathbf{x}_{\text{norm}}, t_{\text{norm}}; \lambda_{\text{norm}}) \\ &\qquad T_{\max}(\lambda) \triangleq \frac{\eta P}{2\pi k r_{\text{laser}}} \end{aligned} \tag{12}$$

This formulation offers two key advantages. First, it significantly reduces inter-material variance in gradient scales by constraining the temperature magnitude with respect to material-dependent thermal properties. Without such a constraint, the optimization process tends to be dominated by materials with excessively large $T_{\max}$ values, as their loss terms induce disproportionately large gradients that bias the training dynamics. Second, the proposed scaling stabilizes the optimization of the PDE, boundary condition, and initial condition losses, all of which are implicitly scaled by $T_{\max}(\lambda)$. While this effect may not be very significant in non-parametric PINNs, it becomes critical in material-agnostic settings, where manual scaling is infeasible or improper scaling can lead to severe gradient imbalance across different materials.

An alternative approach would be to introduce a small, auxiliary fully connected neural network to learn $T_{\max}$ directly (i.e., $g_\theta : \lambda \mapsto \hat{T}_{\max}(\theta)$). However, this strategy would impose additional stochasticity during training and complicate the model architecture. The introduction of the extra learnable components can further exacerbate optimization difficulty and training instability. In contrast, leveraging a physics-based formulation to approximate $T_{\max}$ for each material provides a deterministic, yet effective and principled means of stabilizing the training process of parametric PINNs without increasing architectural complexity.

Here, the correction factor $\kappa$ acts as a hyperparameter that scales the approximated peak temperatures $T_{\max}(\lambda)$ across materials $\lambda$. As illustrated in Fig. 3(b), its fundamental purpose is to map the raw network output $\hat{T}_\Theta$ onto a physically admissible scale $\hat{T}_{\text{phys}}(\Theta)$. In doing so, it prevents the subsequent term $\text{Softplus}(\hat{T}_\Theta)$ from becoming either too small or too large, which would otherwise place an excessive burden on the network $\mathcal{F}_\Theta$ and induce training instability through gradient vanishing or explosion.

The choice of $\kappa$ and its effect can be further understood through how the scaled peak temperature $\kappa T_{\max}(\lambda)$ relates to the true physically admissible temperature range. When $\kappa$ is chosen such that $\kappa T_{\max}(\lambda)$ slightly exceeds this range, covering it with a modest margin, stable convergence can be expected. Conversely, when $\kappa$ is set too low so that $\kappa T_{\max}(\lambda)$ falls below the admissible range, the uncovered portion is passed directly to the network $\mathcal{F}_\Theta$ as an additional burden, leading to significant training instability. When $\kappa$ is set higher than necessary, it still covers the entire admissible range and is substantially less detrimental to training stability than an underestimated value; thus, a conservative overestimation of $\kappa$ is generally preferred. Although $\kappa T_{\max}(\lambda)$ may not be a faithful estimator of the true peak temperature due to the inherent process differences between welding and metal AM, its role is to bound the predicted temperature field within a stable regime, enabling stable training of parametric PINNs even under the large variations in temperature field magnitude induced by diverse thermophysical properties across materials.

Altogether, by embedding physics-informed temperature bounds directly into the output transformation, the proposed approach enables stable and efficient training without introducing additional learnable modules or increasing training complexity. This design choice is therefore essential for achieving effective and scalable material-agnostic temperature prediction within the parametric PINN framework. Note that the proposed strategy is further validated through an ablation study comparing diverse output-scaling mechanisms, as well as a sensitivity analysis of the correction factor $\kappa$.

#### 2.5.2. Hybrid optimization

Training deep learning models, including PINNs, is inherently a non-convex optimization problem, where gradient-based methods remain the gold standard in the literature [47]. In practical deep learning, local minima are often acceptable and comparable to a global optimum, but the presence of numerous saddle points in high-dimensional parameter spaces hinders convergence and induces suboptimal performance [48]. Adaptive gradient-based optimizers such as Adam [49] are widely used to mitigate these challenges by enabling rapid escape from saddle regions [50]. Consequently, many PINN-based studies in metal AM adopt Adam for training [26,37]. The update rule is given by:

$$\Theta_{j+1} \leftarrow \Theta_j - lr_{\text{Adam}} \cdot \text{Adam}\left(\nabla_\Theta \mathcal{L}_B(\Theta_j)\right), \tag{13}$$

where $lr_{\text{Adam}}$ is a learning rate and $\nabla_\Theta \mathcal{L}(\Theta_j)$ is the gradient of the loss function with respect to the model parameter $\Theta$ evaluated on a mini-batch $B \subset \mathcal{D} \times \mathcal{M}$.

Despite its effectiveness, optimization in PINNs is often significantly more challenging than in conventional deep learning due to multi-objective loss formulations and the need to compute higher-order derivatives [51]. Therefore, the loss becomes highly sensitive to minute fluctuations in the network weights $\Theta$. To attenuate the fluctuations, previous studies often employ small learning rates with Adam, which leads to slow convergence and extremely long training horizons, sometimes exceeding 50,000–100,000 epochs [26,33,37].

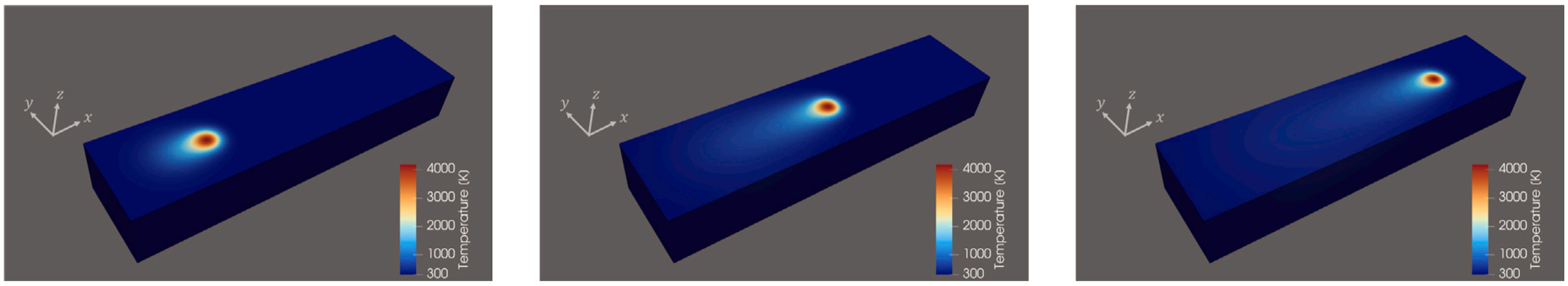

(a) $t = 0.5$ s. (b) $t = 1.5$ s. (c) $t = 2.5$ s.

**Fig. 4.** Ground-truth temperature field snapshots at $t$ = 0.5, 1.5, and 2.5 s for the bare-plate scanning benchmark. Here, we used Ti–6Al–4V for visualization only. The thermal evolution reflects a moving laser heat source and subsequent heat diffusion within the substrate material.

To accelerate convergence, second-order optimizers such as L-BFGS are often used because they exploit curvature information to improve convergence near minima. However, conventional L-BFGS requires full-batch gradients to compute stable curvature estimates. This becomes computationally expensive due to the large number of collocation points. Moreover, the deterministic nature of full-batch updates may lead to optimization stagnation when the training dynamics become trapped in flat regions of the loss landscape.

To address these challenges, we adopt a hybrid optimization strategy that combines Adam for global exploration with L-BFGS for local refinement. The training procedure consists of two phases. First, the model is trained using Adam for $J_{\text{Adam}}$ epochs to efficiently explore the parameter space and obtain a coarse solution. Subsequently, training switches to L-BFGS to achieve precise convergence.

To improve the scalability of L-BFGS, we introduce stochasticity into the collocation sampling. Instead of using all collocation points, fixed subsets of boundary, initial, and PDE residual points are used during each iteration and periodically resampled. This strategy reduces memory consumption while preventing the optimizer from overfitting to specific collocation configurations. The resulting stochastic curvature updates help avoid optimization stagnation and improve training stability. The transition from Adam to L-BFGS can be summarized as

$$\Theta_{j+1} \leftarrow \Theta_j - lr_{\text{L-BFGS}} \cdot \mathbf{H}_j^{-1} \nabla_{\Theta} \mathcal{L}_B(\Theta_J) \quad (14)$$

where $\mathbf{H}_j^{-1}$ denotes the inverse Hessian approximation updated using gradients from a mini-batch $B$ and $lr_{\text{L-BFGS}}$ represents the learning rate.

The overall training procedure of the proposed framework is summarized in Appendix A.2.

## 3. Experiment

*System description.* To validate the proposed framework, we adopt a widely used numerical benchmark [26,33,44] based on a bare-plate laser scanning example reflecting metal AM process. Herein, the high-energy laser selectively melts a metal substrate, producing highly localized melting and rapid thermal transients.

In this benchmark, a moving laser heat source scans the surface of a solid metal plate, generating highly localized and transient thermal gradients governed by laser heating, heat conduction within the substrate, and convective and radiative losses at the free surfaces.

The process parameters are summarized in Table 1. Ground-truth temperature fields are obtained using `JAX-AM`[1] [52], an open-source library for high-fidelity FEM simulations in metal AM. Representative snapshots of the resulting temperature field are shown in Fig. 4.

[1] https://github.com/CMSL-HKUST/jax-am.

**Table 1**
Summary of the experimental setup.

| Process parameter | Value |
|---|---|
| Spatial domain $\Omega$ | $40 \times 10 \times 6$ mm$^3$ |
| Laser power $P$ | 500 W |
| Laser absorptivity $\eta$ | 0.4 |
| Laser beam radius $r_{\text{laser}}$ | 1.5 mm |
| Laser scanning speed $v$ | 10 mm/s |
| Total scanning time $t_{\text{end}}$ | 3 s |
| Initial laser position $\mathbf{x}_0$ | $(5, 5, 6)$ mm |
| Initial temperature $T_0$ | 300 K |
| Ambient temperature $T_\infty$ | 300 K |
| Convection coefficient $h$ | 50 W/(m$^2$ K) |
| Surface emissivity $e$ | 0.3 |

*Material property space.* To evaluate the framework's ability to generalize in capturing material-dependent dynamics, we define a continuous material property space $\mathcal{M}$ based on the characteristic ranges of alloys prevalent in metal AM:

$$\mathcal{M} := [\rho_{\min}, \rho_{\max}] \times [C_{p,\min}, C_{p,\max}] \times [k_{\min}, k_{\max}]$$

where the boundaries are set as $\rho \in [3000, 10\,000]$ kg/m$^3$, $C_p \in [300, 1000]$ J/kg K, and $k \in [3, 50]$ W/m K. Within this space, we validate the model on three representative alloys with material properties denoted as $(\rho, C_p, k)$:

- **Ti-6Al-4V** $(4430, 560, 6.7)$: High strength-to-weight ratio, standard in aerospace.
- **Inconel 718** $(8220, 435, 11.4)$: Nickel-based superalloy with high-temperature stability.
- **SS 316L** $(8000, 500, 16.0)$: Austenitic stainless steel with superior corrosion resistance.

Furthermore, to assess the framework's extrapolation capability in out-of-distribution scenarios, we evaluate it against an additional alloy whose properties significantly exceed the training bounds:

- **AlSi10Mg** $(2670, 950, 150)$: Primarily used for lightweight structural parts, featuring a thermal conductivity $k$ that is nearly threefold the upper limit of the defined $\mathcal{M}$.
- **Copper** $(8960, 385, 401)$: Highly conductive material widely used in electrical and thermal management components, exhibiting an extreme thermal conductivity $k$ that exceeds the upper bound of $\mathcal{M}$ by eightfold.

*Governing equations and boundary conditions.* We leave the detailed mathematical formulation of the governing PDEs as well as the boundary conditions and initial conditions to Appendix A.3.

*Collocation sampling strategy.* To train PINN models using physics-based residuals, collocation points sampled from the spatiotemporal domain are employed. Although the choice of sampling strategy is critical and constitutes a broad research topic in itself, we follow the established setup from prior work [26] to ensure consistency and enable fair comparison. Under this configuration, the sampled collocation points used throughout this study are illustrated in Fig. 5.

**Table 2**
Summary of network architectures, inputs, and training configurations for the proposed model and the baselines.

| Model | Input | Architecture | Layers/Width | Optimizer/Learning rate/Epochs |
|---|---|---|---|---|
| N-PINN [26] | $(x, y, z, t)$ | Monolithic | 4/60 | Adam/2e−4/50,000 |
| P-PINN | $(x, y, z, t, \rho, C_p, k)$ | Monolithic | 4/60 | Hybrid/2e−4, 1e−3/10,000 |
| Proposed | $(x, y, z, t, \rho, C_p, k)$ | Decoupled | $(2/30) \times 2 + (2/50)$ | Hybrid/2e−4, 1e−3/10,000 |

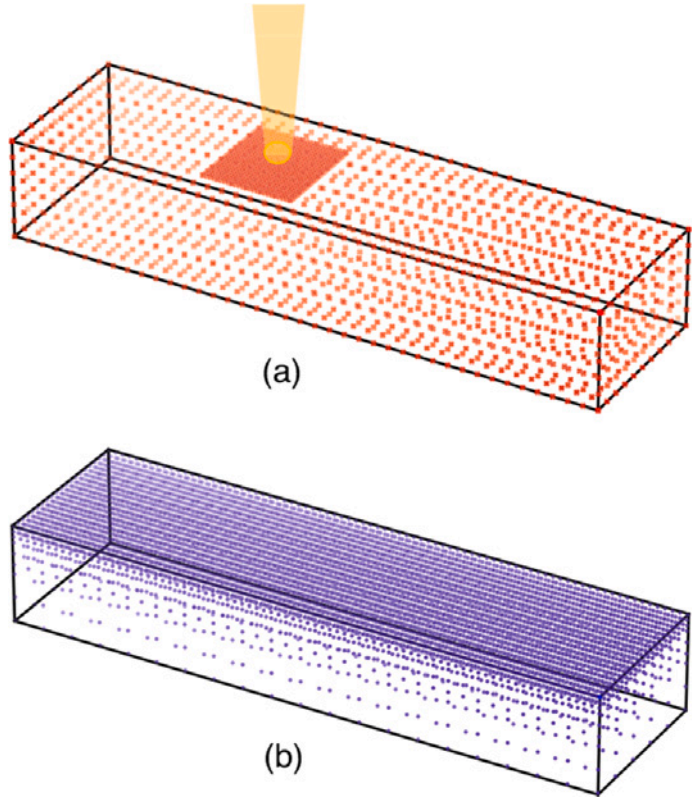

**Fig. 5.** Visualization of the sampled collocation points used in this study. Sampling points at time $t$ (a) on the boundary (b) in the domain.
*Source:* Reproduced from [26].

Specifically, the spatiotemporal domain is discretized with a uniform time step of 0.1 s. At each step, a multi-resolution sampling strategy is applied: a baseline 1 mm interval grid is used for the surfaces, while a refined 0.25 mm interval is concentrated within a $6 \times 6$ mm$^2$ area near the laser center. Additionally, a 0.5 mm grid is utilized for the top $40 \times 10 \times 1$ mm$^3$ region — four times finer than the lower region — to ensure that high-frequency thermal behaviors and transient boundary conditions near the heat source are accurately captured.

*Baselines and evaluation metric.* We evaluate the proposed material-agnostic framework against two baseline models to assess the scalability and architectural advantages of our approach.

- **Non-parametric PINN (N-PINN):** Material properties are treated as fixed constants within the governing equations. This baseline assesses whether the material-agnostic framework preserves predictive accuracy while providing improved scalability across different materials.
- **Monolithic parametric PINN (P-PINN):** A conventional parametric PINN architecture that takes all physical and material inputs as a single vector (e.g., Fig. 2). This comparison evaluates the architectural benefits of the proposed decoupled design relative to the commonly used parametric PINN structure in metal AM studies [36,37].

For quantitative comparison, we adopt the relative $L_2$ error, a standard metric in the PINN literature. This metric measures the normalized discrepancy between the predicted temperature field and the ground-truth solution over the discretized spatiotemporal domain. For a given set of material properties $\lambda$, the relative $L_2$ error is defined as:

$$L_2 \text{ error} = \frac{\sqrt{\sum_{n=1}^{N} \left| T\left(\mathbf{x}^{(n)}, t^{(n)}, \lambda\right) - \hat{T}_{\text{phys}}\left(\mathbf{x}^{(n)}, t^{(n)}, \lambda; \Theta\right) \right|^2}}{\sqrt{\sum_{n=1}^{N} \left| T\left(\mathbf{x}^{(n)}, t^{(n)}, \lambda\right) \right|^2}} \times 100\,(\%)$$

where $N$ denotes the total number of nodes in the FEM simulations. In this study, $N = 88{,}939$.

**Table 3**
Quantitative performance comparison with the N-PINN baseline [26]. We report the total number of learnable parameters and the $L_2$ error [%] over five random seeds as mean ± standard deviation. All experimental settings were identical except for the number of training epochs to ensure fair comparison. Lower values indicate better performance and parameter efficiency.

| Material $(\rho, C_p, k)$ | # param. | Ti–6Al–4V $(4430, 560, 6.7)$ | Inconel 718 $(8220, 435, 11.4)$ | SS 316L $(8000, 500, 16)$ |
|---|---|---|---|---|
| N-PINN [26] | 11,341 | $6.19 \pm 1.00$ | $6.09 \pm 0.83$ | $5.94 \pm 0.63$ |
| **Proposed** | **9641** | $\mathbf{2.71} \pm 0.34$ | $\mathbf{2.18} \pm 0.45$ | $\mathbf{2.17} \pm 0.53$ |
| Reduction [%] | **−14.99%** | **−56.22%** | **−64.20%** | **−63.43%** |

*Implementation details.* For the N-PINN baseline, we adopt the implementation provided in the corresponding study [26]. The network architectures, input variables, and training configurations for the proposed model and the baseline methods are summarized in Table 2. It is worth noting that the proposed model has the fewest trainable parameters while achieving superior predictive performance, as demonstrated in the following section.

Aside from these configurations, all experimental settings were identical, and all experiments were conducted using five random seeds to ensure reproducibility and fair comparison. The loss-function weights were set to $(w_{\text{pde}}, w_{\text{bc}}, w_{\text{ic}}) = (1, 1, 10^{-4})$ following prior work [26]. Detailed weight analysis is in Appendix A.5. A curriculum learning strategy was applied during the first 200 epochs to prevent convergence to trivial local optima by prioritizing the satisfaction of boundary conditions (i.e., $(w_{\text{pde}}, w_{\text{bc}}, w_{\text{ic}}) = (0, 1, 0)$). For the proposed hybrid optimization strategy, we transition from Adam to L-BFGS at 2000 epochs. Additional implementation details, computation time, and hardware used in this study are provided in Appendix A.4. The complete implementation and data are available at https://github.com/hsleecri/MaterialAgnosticTempPred.

## 4. Results and discussion

### 4.1. Comparison with non-parametric PINN

Table 3 summarizes the $L_2$ error for three alloys: Ti–6Al–4V, Inconel 718, and SS 316L. The proposed framework consistently outperforms the N-PINN baseline across all materials, achieving approximately 56–64% lower $L_2$ error while using fewer trainable parameters. This result is particularly noteworthy given that the N-PINN baseline requires material-specific retraining, whereas the proposed framework enables simultaneous evaluation across all materials within a single trained model while maintaining strong cross-material generalization.

Beyond final accuracy, Fig. 6(a)–(c) highlights a substantial difference in training behavior. Notably, after switching from Adam to L-BFGS at 2000 epochs, only an additional 100–200 epochs are required for the proposed model to reach the performance level achieved by the N-PINN baseline after approximately 50,000 epochs of training. In other words, the proposed framework attains a comparable accuracy within 2200 epochs, which is 4.4% of the baseline training iterations conducted in the corresponding previous study.

With continued training, the performance gap further widens. At 10,000 epochs, the proposed framework achieves an additional 56–64% reduction in $L_2$ error relative to the N-PINN, indicating that the improvement extends beyond faster convergence to a superior final solution. This reduction in required training iterations is particularly significant for PINNs in metal AM, where prior studies often rely on extremely long training horizons to achieve satisfactory performance. In contrast, the proposed framework demonstrates that rapid convergence and improved final accuracy can be achieved simultaneously without excessive training.

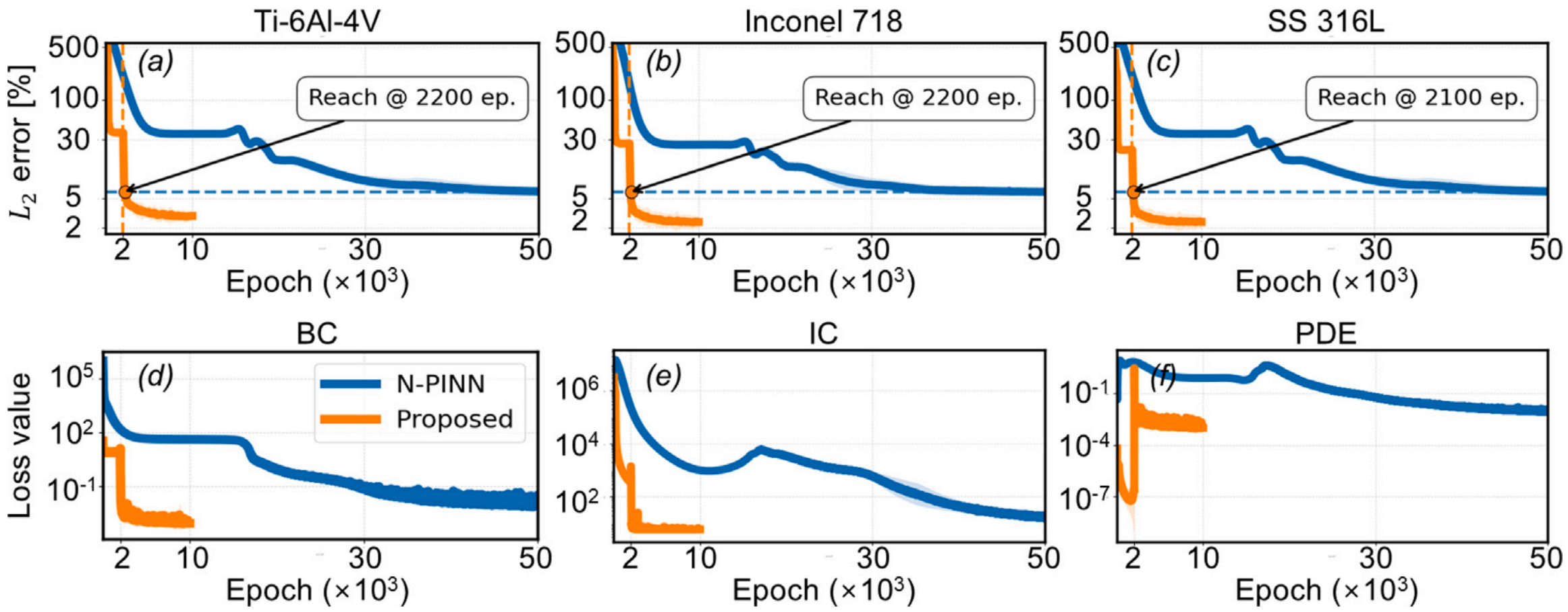

**Fig. 6.** Comparison of training dynamics between the N-PINN [26] and the proposed framework. (a)–(c) Evolution of the $L_2$ error [%] during training for Ti–6Al–4V, Inconel 718, and SS 316L. (d)–(f) Evolution of the loss components in (2): boundary condition loss (4), initial condition loss (5), and PDE residual loss (3). For loss visualization, Ti–6Al–4V is shown for the N-PINN. Solid lines represent the mean over five random seeds, and shaded regions indicate ± one standard deviation.

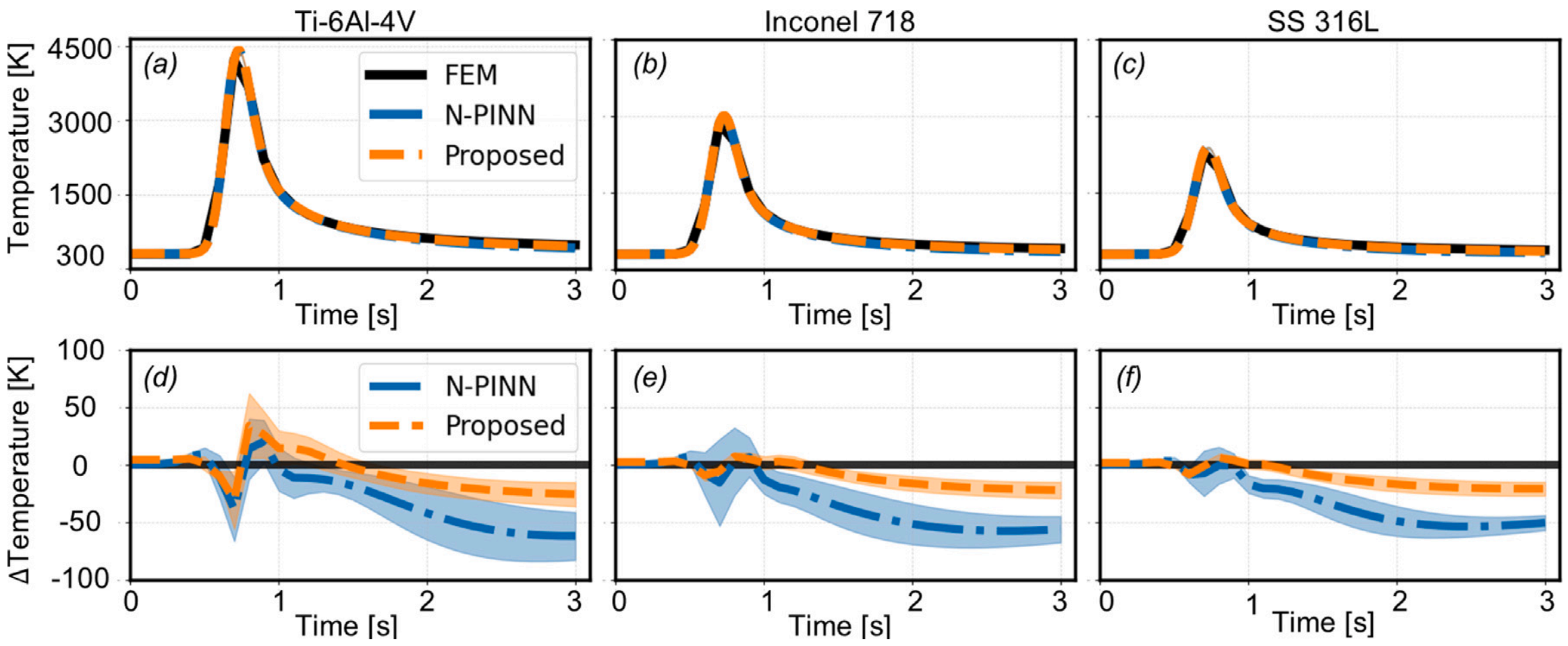

**Fig. 7.** Thermal history comparison at a probe location along the laser scan path on the top surface for Ti–6Al–4V, Inconel 718, and SS 316L. (a)–(c) Temperature evolution predicted by FEM, the N-PINN [26], and the proposed framework. (d)–(f) Temperature deviation relative to the FEM reference, defined as $\Delta T = \hat{T}_{\mathrm{phys}}(\Theta) - T_{\mathrm{FEM}}$. Solid lines represent the mean over five random seeds, and shaded regions indicate ± one standard deviation.

To investigate the origin of this performance gap, Fig. 6(d)–(f) compares the evolution of individual PINN loss components. After transitioning to L-BFGS at 2000 epochs, the proposed framework exhibits a rapid and stable decrease in all loss terms. While the initial condition loss converges to a similar lower bound for both methods, notable differences arise in the boundary condition and PDE residual losses. In particular, the proposed framework reduces the loss values by up to $1/100$ compared to the N-PINN while maintaining stable optimization. This substantial reduction in physics-based residuals directly explains the improved accuracy reported in Table 3.

The impact of these improvements is further scrutinized in Fig. 7(a)–(c), which compares thermal histories evaluated at a probe location $(12, 5, 6)$ on the top surface along the laser scan path. As the laser approaches the probe location, the temperature signal enters a highly transient regime characterized by steep thermal gradients. This regime is particularly challenging for PINN models, since neural networks exhibit a spectral bias toward low-frequency functions and therefore struggle to accurately represent the sharp thermal variations induced by localized laser heating. During this phase, both the proposed framework and the N-PINN baseline capture the peak temperature induced by laser irradiation and reproduce the distinct thermal scales associated with different materials.

More pronounced differences emerge during the subsequent cooling phase after the laser passes the probe location. The N-PINN exhibits systematic bias and larger variability, consistent with residual boundary condition and PDE errors. In contrast, the proposed framework remains closer to the FEM reference solution, yielding smaller deviations and reduced variance, particularly in cooling-dominated regimes. These results indicate that the reduced boundary condition and PDE residuals translate directly into more stable and accurate material-agnostic thermal predictions.

An additional trend is observed across the three alloys. The largest prediction errors occur for Ti–6Al–4V, which has relatively low thermal conductivity and therefore exhibits sharp and highly localized thermal gradients near the melt pool region. Such high-gradient regimes are difficult for PINNs to approximate accurately and may also be insufficiently resolved by the adopted collocation sampling density. As thermal conductivity increases from Ti–6Al–4V to Inconel 718 and SS 316L, heat diffusion becomes more dominant, smoothing sharp temperature gradients and attenuating localized thermal fluctuations. Consequently, the prediction errors near the melt pool region decrease for both models.

**Table 4**
Quantitative performance comparison between the monolithic parametric PINN (P-PINN) baseline and the proposed decoupled framework. We report the total number of learnable parameters and the relative $L_2$ error [%] over five random seeds as mean ± standard deviation, with the observed range [min, max] shown in each entry. All experimental settings were identical except for the number of training epochs to ensure fair comparison. Lower values indicate better parameter efficiency and performance.

| Material $(\rho, C_p, k)$ | # param. | Ti–6Al–4V (4430, 560, 6.7) | Inconel 718 (8220, 435, 11.4) | SS 316L (8000, 500, 16) |
|---|---|---|---|---|
| Parametric PINN (P-PINN) | 11,521 | $7.59 \pm 3.86$ [4.10, 13.35] | $6.12 \pm 3.22$ [3.39, 11.14] | $5.84 \pm 2.59$ [3.63, 9.90] |
| **Proposed** | **9641** | $\mathbf{2.71} \pm 0.34$ [2.35, 3.03] | $\mathbf{2.18} \pm 0.45$ [1.76, 2.69] | $\mathbf{2.17} \pm 0.53$ [1.67, 2.11] |
| Reduction [%] | **−16.32%** | **−64.30%** | **−64.38%** | **−62.84%** |

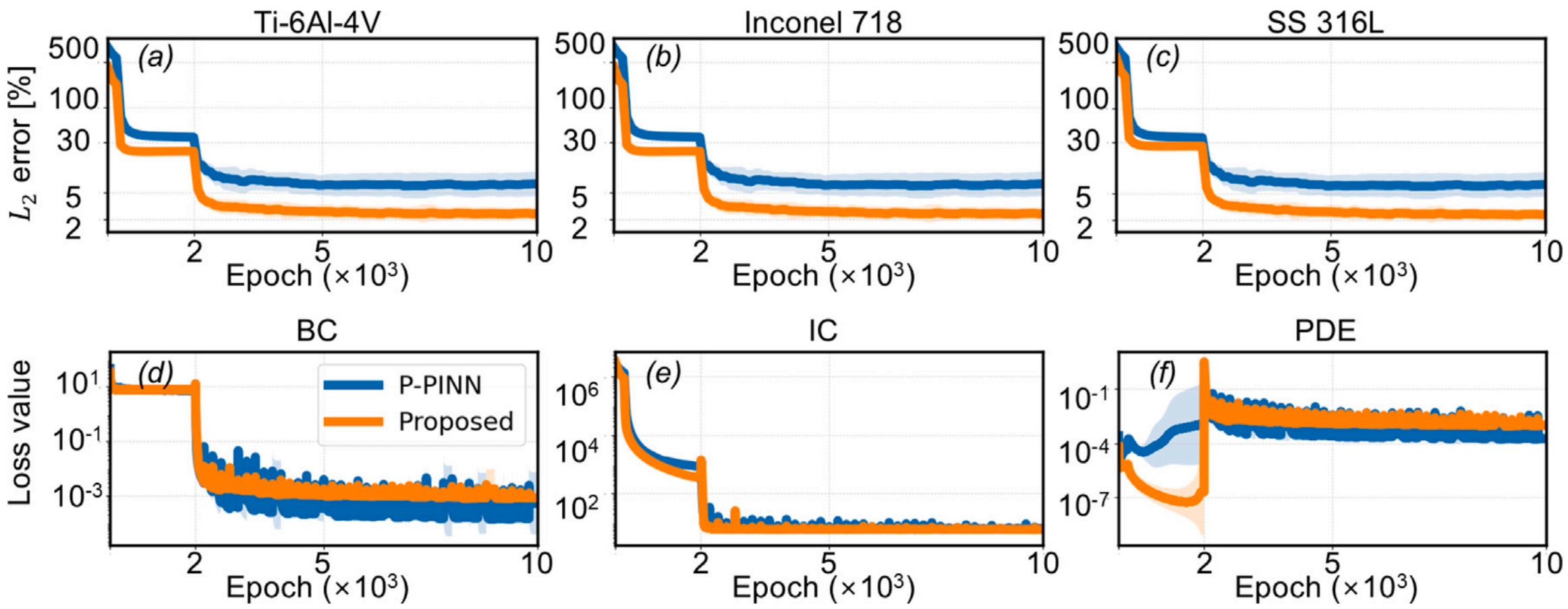

**Fig. 8.** Comparison of training dynamics between the P-PINN and the proposed framework. (a)–(c) Evolution of the $L_2$ error [%] during training for Ti–6Al–4V, Inconel 718, and SS 316L. (d)–(f) Evolution of the loss components in (2): boundary condition loss (4), initial condition loss (5), and PDE residual loss (3). Solid lines represent the mean over five random seeds, and shaded regions indicate ± one standard deviation.

Despite these improvements for higher-conductivity materials, the proposed framework consistently shows smaller deviations and lower variance than the N-PINN in the late-time far-field regime across all three alloys. This behavior suggests that the N-PINN tends to learn material-specific thermal responses due to its non-parametric formulation, which can lead to material-specific overfitting. In contrast, the proposed parametric architecture learns shared thermal dynamics across multiple materials, enabling more generalizable and stable temperature predictions.

### 4.2. Comparison with monolithic parametric PINN

Table 4 summarizes the quantitative results in terms of the total number of learnable parameters and the relative $L_2$ error. Despite having 16.32% fewer parameters, the proposed framework consistently improves predictive accuracy compared to the P-PINN baseline across all materials, achieving relative $L_2$ error reductions of 64.3%, 64.38%, and 62.84% for Ti–6Al–4V, Inconel 718, and SS 316L, respectively, while also exhibiting substantially smaller standard deviations. Note that the only difference between the two models lies in the parametric PINN architecture, while all other experimental settings such as training horizons are identical.

A closer inspection of the minimum and maximum $L_2$ error values reveals a pronounced contrast in robustness. For all three materials, even the best-performing instances of the P-PINN (i.e., the minimum error across random seeds) remain worse than the worst-case performance of the proposed framework. Beyond this instance-level comparison, the proposed framework maintains a consistently narrow performance range across all materials, whereas the P-PINN exhibits large variance due to training instability. This behavior suggests that the decoupled architecture stabilizes the learning process by mitigating gradient interference and enabling more effective representation of the interactions between spatiotemporal features and material parameters, resulting in improved alignment with the underlying physics.

The training dynamics are further analyzed in Fig. 8. Both the P-PINN and the proposed framework exhibit smooth and stable convergence behavior, with all loss components consistently decreasing across random seeds. Unlike the pronounced discrepancies observed between the N-PINN and the proposed method, all physics-based residuals for P-PINN and the proposed framework converge to comparable magnitudes, indicating that both models are effectively optimized and guided toward similar regions of the loss landscape by the hybrid optimization strategy.

Despite this similarity in loss convergence, the P-PINN consistently yields higher $L_2$ errors than the proposed framework. This indicates that the remaining performance gap is not primarily driven by optimization failure, but rather by limitations in representation. In particular, the monolithic P-PINN must encode spatiotemporal coordinates and material properties within a shared latent representation via additive feature fusion, which constrains its ability to efficiently capture the inherently multiplicative role of material parameters in the governing equations.

In contrast, the proposed decoupled architecture enables material properties to act as conditional modulation variables that multiplicatively scale and shift spatiotemporal features, yielding representations that are better aligned with the physical structure of the PDEs and boundary conditions. This improved representational alignment allows

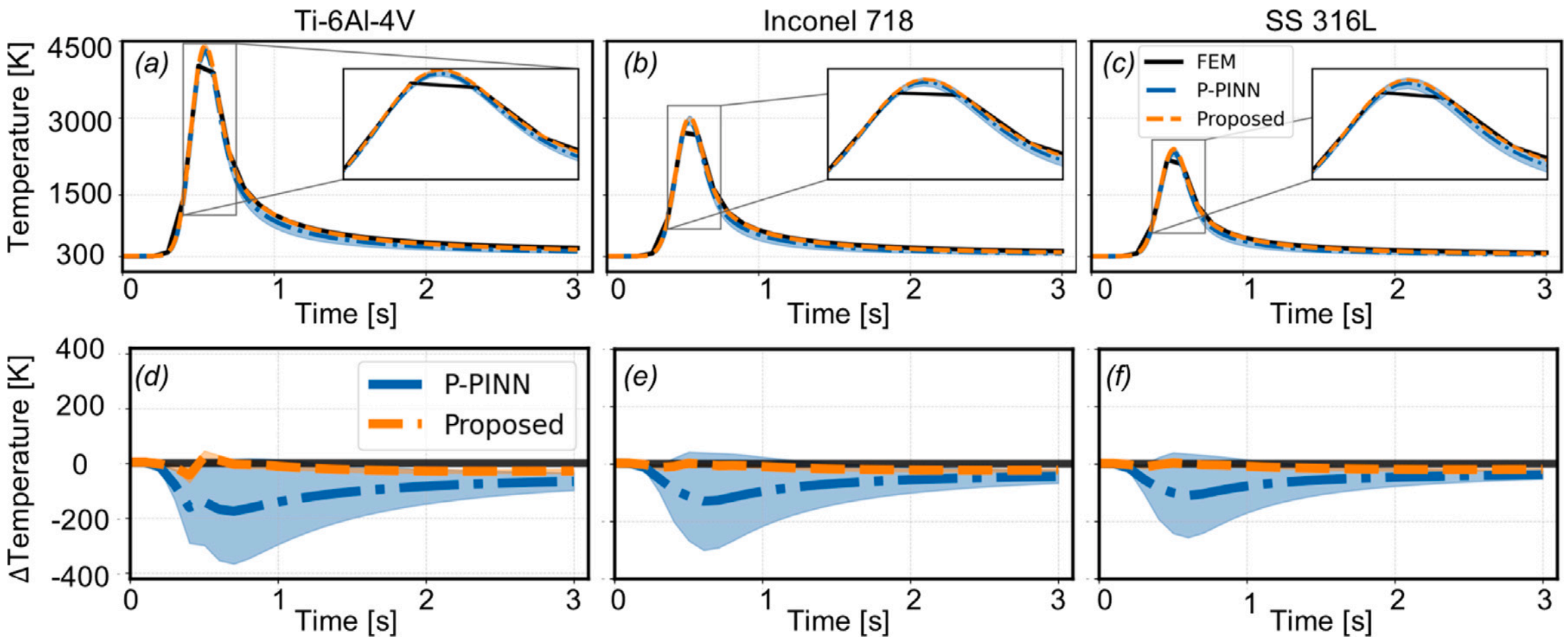

**Fig. 9.** Thermal history comparison at a probe location along the laser scan path on the top surface for Ti–6Al–4V, Inconel 718, and SS 316L. (a)–(c) Temperature evolution predicted by FEM, the P-PINN, and the proposed framework. (d)–(f) Temperature deviation relative to the FEM reference, defined as $\Delta T = \hat{T}(\Theta) - T_{\text{FEM}}$. Solid lines represent the mean over five random seeds, and shaded regions indicate ± one standard deviation.

comparable physics-based loss minimization to translate into superior predictive accuracy.

Fig. 9(a)–(c) compares the thermal histories at a probe location $(9, 5, 6)$ on the top surface along the laser scan path. Across all materials, the proposed framework closely follows the FEM reference throughout both the heating and cooling phases. In contrast, the P-PINN exhibits noticeable deviations near the peak temperature and during the subsequent transient cooling regime, accompanied by substantially larger uncertainty bands. Notably, the largest deviations for the P-PINN occur around the peak temperature induced by laser irradiation. This behavior suggests that, while some P-PINN instances converge to adequate local optima that capture the rapid thermal response to laser heating, others converge to solutions that do not effectively represent the underlying thermal dynamics. Such sensitivity to initialization and optimization leads to significant variability.

Conversely, the proposed framework consistently captures the laser-induced heating and subsequent cooling behavior, indicating that its parameter-efficient and decoupled architecture provides representations that are better aligned with the governing equations and thermal physics of metal AM. This observation further supports the claim that architectural decoupling improves the expressiveness of the model in capturing the interaction between material properties and spatiotemporal thermal dynamics.

### 4.3. Material-agnostic thermal field prediction

Beyond the quantitative evaluation, we qualitatively examine the spatiotemporal temperature fields predicted by the proposed framework, focusing on their temporal evolution and material-dependent behavior. As in the quantitative experiments, the model is trained only once and evaluated across different materials.

At the early stage ($t = 0.5$ s; Fig. 10), the temperature field is dominated by highly localized peaks near the moving laser heat source. Both the top-surface distributions and the $x$–$z$ cross-sections predicted by the proposed framework closely match the FEM reference in terms of peak location, spatial extent, and symmetry. The residual error $\Delta T$ remains strongly localized around the melt pool region, indicating that the dominant conductive heat-transfer mechanisms are accurately captured at early times.

As the thermal field evolves ($t = 1.5$ s; Fig. 11), heat accumulates along the scanning direction, producing an elongated temperature tail behind the moving heat source and deeper heat penetration below the surface. Although the magnitude of $\Delta T$ increases slightly compared with the early-time case, the error remains localized near the heat source and along the trailing region, suggesting that the model preserves the correct heat-transport behavior without accumulating large errors over time.

At the later stage ($t = 2.5$ s; Fig. 12), heat accumulation becomes more pronounced, particularly in subsurface regions where conduction dominates. Despite a modest increase in the spatial extent of $\Delta T$, the predicted temperature fields remain in strong qualitative agreement with the FEM solutions across all materials. No evidence of global drift or temporal instability is observed, indicating stable long-term behavior even when the temperature field is strongly influenced by prior heating.

Across all time instants, the residual error patterns exhibit consistent spatial structures among different materials despite significant differences in temperature levels and thermophysical properties. Residual discrepancies are primarily concentrated near the melt pool region, where steep temperature gradients and rapidly evolving thermal dynamics produce high-frequency spatiotemporal features that are inherently challenging for PINN models to capture. Among the materials considered, Ti–6Al–4V shows slightly larger discrepancies near the melt pool due to its lower thermal diffusivity and steeper temperature gradients, whereas Inconel 718 and SS 316L exhibit smoother thermal fields and correspondingly smaller residual magnitudes. These observations are consistent with the probe-based thermal history analysis discussed earlier.

Notably, the spatial distribution of errors varies systematically with the thermal conductivity $k$. For materials with lower thermal conductivity, the largest discrepancies appear near the melt pool region, where steep gradients generate high-frequency thermal features that are difficult to resolve. In contrast, materials with higher thermal conductivity exhibit smoother temperature fields due to stronger diffusion, which alleviates errors near the melt pool but introduces larger discrepancies in far-field and late-time regions where accumulated heat diffusion dominates the thermal response.

These behaviors may, in part, be attributed to the identical collocation sampling strategy adopted for all materials in this study. Because different materials exhibit distinct thermal diffusion rates and gradient magnitudes depending on their thermophysical properties, a uniform sampling strategy may be insufficient to resolve the material-dependent spatiotemporal scales present in the temperature field. This limitation suggests that adaptive or material-conditioned sampling strategies could further improve the accuracy of PINN-based thermal models by better resolving both high-frequency melt pool dynamics and diffusion-dominated far-field regimes.

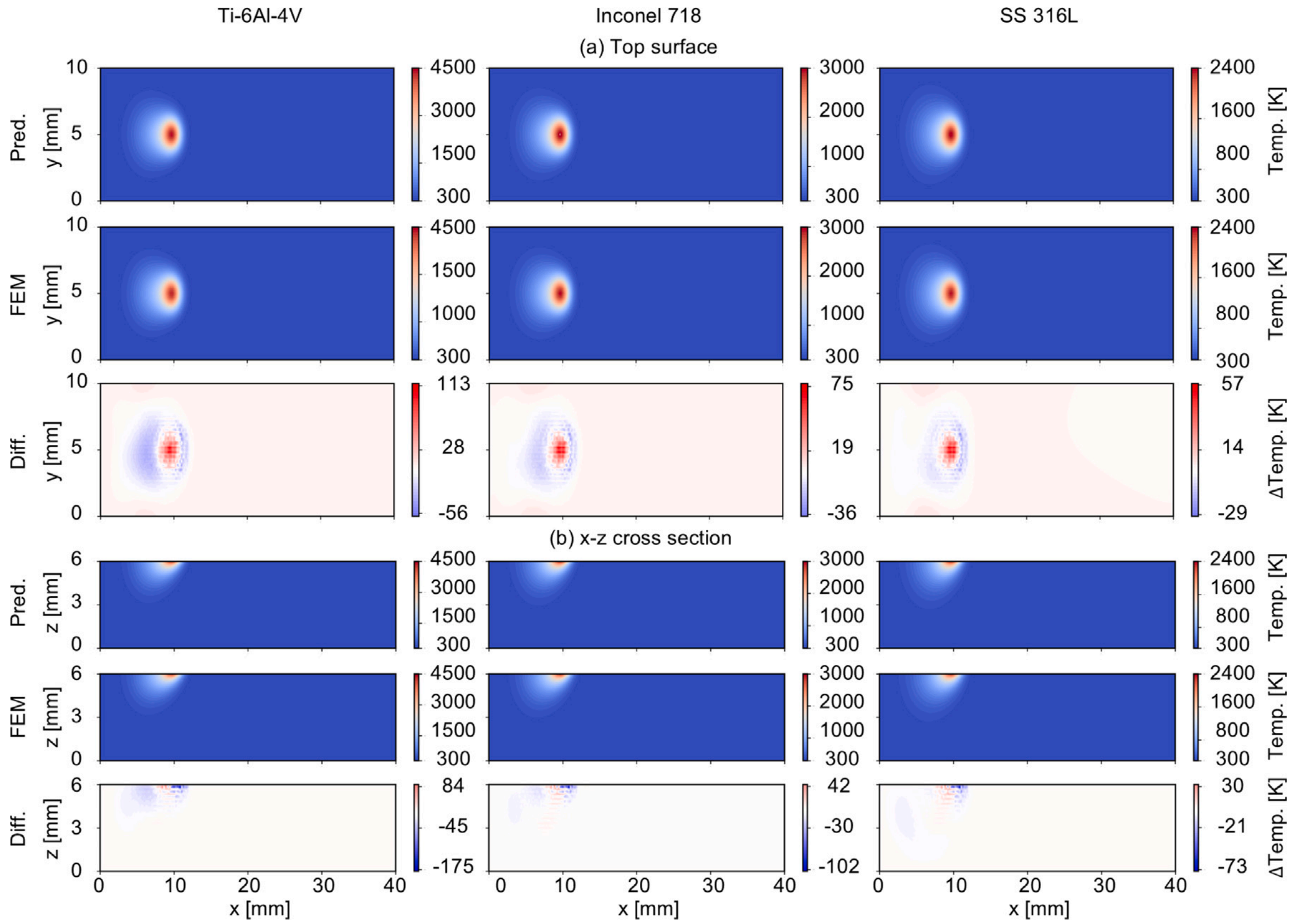

**Fig. 10.** Qualitative comparison of temperature fields at $t = 0.5$ s predicted by the proposed framework and FEM simulations for three materials: Ti–6Al–4V, Inconel 718, and SS 316L. (a) Top-surface temperature fields predicted by the proposed framework, the corresponding FEM reference solutions, and the discrepancy $\Delta T = \hat{T}_{\text{phys}}(\Theta) - T_{\text{FEM}}$. (b) the corresponding $x$–$z$ cross-sections following the same ordering.

**Table 5**
Quantitative performance comparison with baselines for OOD scenarios: AlSi10Mg and Copper. We report the relative $L_2$ error [%] over five random seeds as mean ± standard deviation. All experimental settings were identical except for the number of training epochs to ensure fair comparison. Lower values indicate better performance.

| Material $(\rho, C_p, k)$ | AlSi10Mg (2670, 950, 150) | Copper (8960, 385, 401) |
|---|---|---|
| N-PINN [26] | $4.25 \pm 0.35$ | $1.63 \pm 0.13$ |
| P-PINN | $3.25 \pm 0.39$ | $14.38 \pm 23.92$ |
| **Proposed** | **1.75** $\pm 0.28$ | **0.69** $\pm 0.14$ |

### 4.4. *Temperature prediction on the materials beyond the defined material property space*

We further evaluate the proposed framework's generalizability for materials beyond the defined material property space $\mathcal{M}$. In the ML literature, generalization is typically categorized into in-distribution (ID) and out-of-distribution (OOD) settings. In our case, the distribution is defined over material properties $\lambda = (\rho, C_p, k) \in \mathcal{M}$. It is worth noting that generalizing to OOD scenarios is widely considered substantially more challenging than generalizing within the training distribution. Accordingly, we select AlSi10Mg and Copper as OOD test cases, with material properties $(\rho, C_p, k) = (2670, 950, 150)$ and $(8960, 385, 401)$, respectively. In both cases, their thermal conductivity values, $k = 150$ and $401$ W/(m K), fall well outside the training range of $k \in [3, 50]$ W/(m K).

Table 5 demonstrates that the proposed framework outperforms all baseline methods across both OOD test cases, AlSi10Mg and Copper. For AlSi10Mg, all baseline methods achieve reasonably accurate predictions with stable performance. For Copper, however, P-PINN exhibits substantially higher error and variance, indicating a clear performance breakdown.

A plausible explanation lies in the significant difference in thermal diffusivity $\alpha$ between Copper and the training material property space. For AlSi10Mg, the thermal diffusivity is $\alpha_{\text{Al}} = 59.14$ mm$^2$/s, while the training material property space $\mathcal{M}$ spans $\alpha \in [0.3, 55.56]$ mm$^2$/s. Although AlSi10Mg lies slightly outside the training range, its diffusivity is close to the upper bound of the training domain, which may explain why P-PINN retains reasonable performance in this case.

In contrast, Copper exhibits a substantially larger thermal diffusivity of $\alpha_{\text{Cu}} = 116.24$ mm$^2$/s, far exceeding the range of all training materials. This represents a considerably more challenging OOD scenario and likely accounts for the observed performance degradation of P-PINN. The proposed framework, by comparison, maintains accurate thermal predictions even under this extreme condition, achieving an $L_2$ relative error below 1%.

Fig. 13 presents qualitative comparisons with the FEM reference solution at $t \in \{0.5, 1.5, 2.5\}$ s for Copper. The proposed framework closely matches the FEM results across all time snapshots, achieving an $L_2$ error below 1% while maintaining a maximum absolute temperature deviation of less than 10 K.

In addition, these qualitative results suggest a promising direction for future work: Collocation point sampling strategies could be adapted based on material thermophysical properties. For instance, concentrating collocation points near high-gradient regions around the laser footprint may improve resolution of rapid boundary transients for low-conductivity materials. For high-conductivity materials, the dominant thermal evolution tends to occur in late-time and far-field regions with smoother temperature gradients; allocating more collocation points in these regions could therefore enhance modeling accuracy through material-aware sampling.

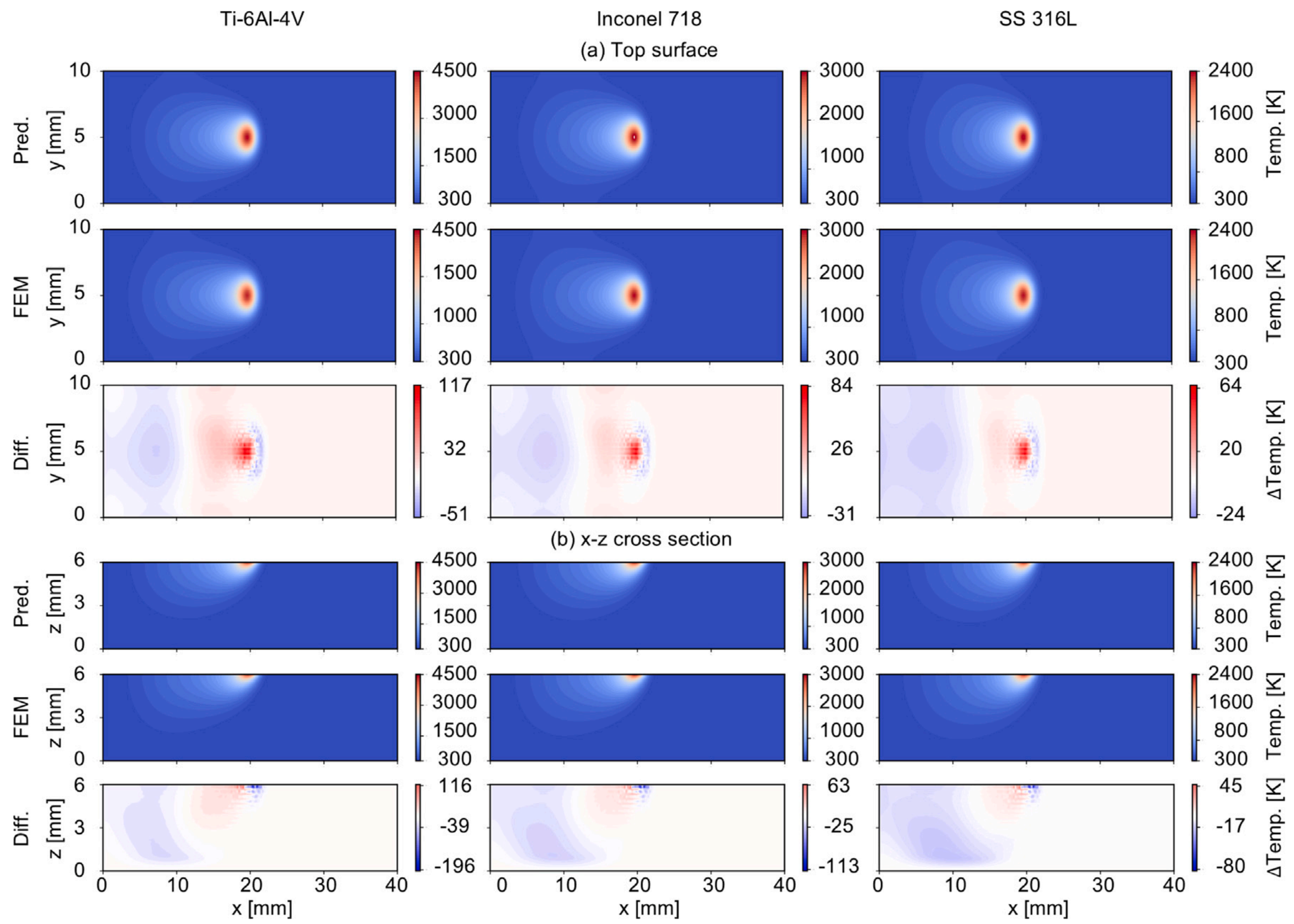

**Fig. 11.** Qualitative comparison of temperature fields at $t = 1.5$ s predicted by the proposed framework and FEM simulations for three materials: Ti–6Al–4V, Inconel 718, and SS 316L. (a) Top-surface temperature fields predicted by the proposed framework, the corresponding FEM reference solutions, and the discrepancy $\Delta T = \hat{T}_{\text{phys}}(\Theta) - T_{\text{FEM}}$. (b) the corresponding $x$–$z$ cross-sections following the same ordering.

### 4.5. Validation under different process parameters settings

To further validate our framework under different process parameter settings, we varied the laser power and scanning speed by ±10%, as these two parameters are typically the most significant factors in metal AM. Our parameter variations follow the experimental setup established in previous work [33]. Specifically, we introduced laser power $P \in \{450, 500, 550\}$ and scanning speed $v \in \{9, 10, 11\}$, whereas our baseline setup was $P = 500$ and $v = 10$. Accordingly, only the laser paths change in response to the scanning speed variations, while other experimental settings (e.g., the total scanning time $t_{\text{end}}$ and the collocation sampling strategy) remain identical.

Note that models are trained separately for each process parameter configuration, as the aim and scope of this work are cross-material generalization. Extending the framework to jointly handle unseen process conditions including both material and process parameter changes within a single trained model is left for our future work.

Table 6 reports the overall performance across the diverse process parameter settings. Beyond the baseline experimental setup ($P = 500\,\text{W}$, $v = 10\,\text{mm/s}$), the proposed framework demonstrates consistently robust performance on the eight additional process conditions, across all five materials and spanning both in-distribution and OOD cases. We attribute this strong performance to the key components of our framework, particularly the physics-guided output scaling grounded in the Rosenthal analytical solution.

Specifically, even when changes in process parameters affect the overall thermal behavior, especially the peak temperatures, our physics-guided output scaling effectively tracks the trend of these peak temperature variations. To quantify this, let $\bar{\gamma}$ denote the mean ratio of our approximated peak temperatures $T_{\max}(\lambda)$ to the peak temperatures obtained from the high-fidelity FEM simulation $T_{\max,\text{FEM}}(\lambda)$ (i.e., $\bar{\gamma} = \text{mean}\left(\frac{T_{\max}(\lambda)}{T_{\max,\text{FEM}}(\lambda)}\right)$) across the five evaluation materials.

As shown in Fig. 14, we observed that $\bar{\gamma}$ generally remained around 0.74 due to the inherent differences between metal AM and the welding processes underlying the Rosenthal analytical solution. To narrow this gap, we introduced the hyperparameter $\kappa$ and set it to 1.5, scaling up $T_{\max}(\lambda)$ and thereby increasing $\bar{\gamma}$ to approximately 1.11. This adjustment ensures that our framework safely bounds the physically admissible temperature range. A more detailed analysis of the impact of $\kappa$ is provided in Section 5.1.

### 4.6. Overall discussion

In summary, extensive experiments demonstrate that the proposed framework exhibits strong cross-material generalization across both ID and OOD regimes. Compared to previous PINN approaches, including N-PINN and P-PINN, our framework significantly outperforms the baselines in both performance and efficiency, reducing the $L_2$ error by up to 64% while surpassing their accuracy within only 4.4% of their training horizons with fewer learnable parameters.

These gains stem from our framework's ability to prevent material-specific overfitting. While N-PINN suffers from notable deviations — particularly in the far-field, late-time temperature fields (Fig. 7) — our parametric formulation effectively captures diverse, material-dependent thermal behaviors. Furthermore, unlike conventional monolithic parametric PINNs (P-PINN), which often exhibit severe training instability and fail to consistently outperform non-parametric baselines (Table 4), our framework maintains robust cross-material generalization even in challenging OOD regimes (Table 5). This superior performance is driven by enhanced representation learning via the physics-inductive bias embedded in our architecture, aligning seamlessly with the physics of metal AM.

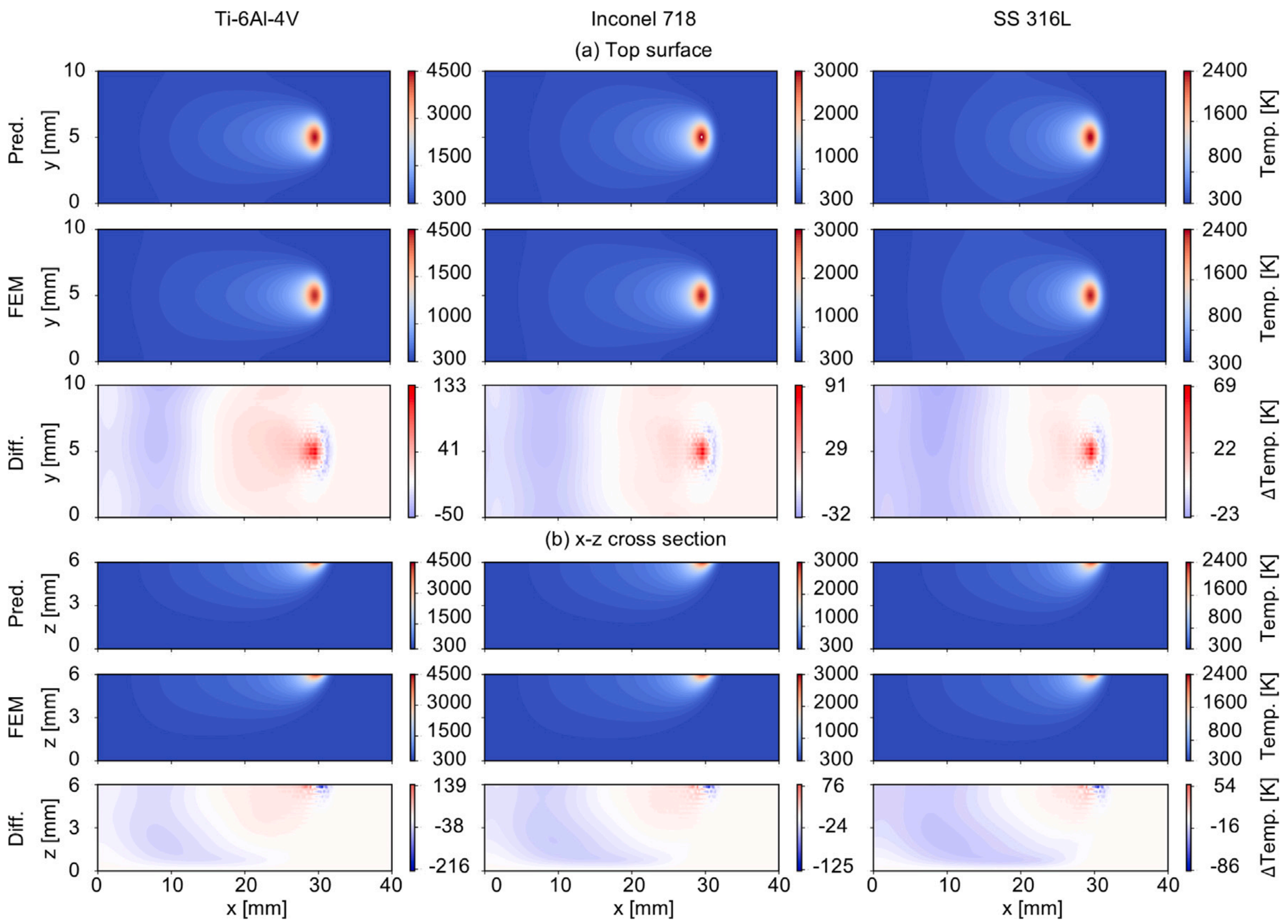

**Fig. 12.** Qualitative comparison of temperature fields at $t$ = 2.5 s predicted by the proposed framework and FEM simulations for three materials: Ti–6Al–4V, Inconel 718, and SS 316L. (a) Top-surface temperature fields predicted by the proposed framework, the corresponding FEM reference solutions, and the discrepancy $\Delta T = \hat{T}_{\mathrm{phys}}(\Theta) - T_{\mathrm{FEM}}$. (b) the corresponding $x$–$z$ cross-sections following the same ordering.

**Table 6**
Per-material $L_2$ error (%, mean ± std over five seeds) of the proposed framework across nine process conditions and five materials: Ti-6Al-4V, Inconel 718, and SS316L are ID, and AlSi10Mg and Copper are OOD. *Average* denotes the mean across all five materials. Material properties are listed as ($\rho$, $C_p$, $k$) in (kg m$^{-3}$, J kg$^{-1}$ K$^{-1}$, W m$^{-1}$ K$^{-1}$).

| Laser power $P$ | Scan speed $v$ | Average | In-distribution (ID) | | | Out-of-distribution (OOD) | |
|---|---|---|---|---|---|---|---|
| | | | Ti-6Al-4V (4430, 560, 6.7) | Inconel 718 (8220, 435, 11.4) | SS316L (8000, 500, 16.0) | AlSi10Mg (2670, 950, 150) | Copper (8960, 385, 401) |
| 450 W | 9 mm/s | 2.77 ± 0.65 | 4.20 ± 0.97 | 3.41 ± 0.82 | 3.39 ± 0.83 | 2.00 ± 0.42 | 0.83 ± 0.18 |
| | 10 mm/s | 1.52 ± 0.21 | 2.08 ± 0.16 | 1.68 ± 0.18 | 1.62 ± 0.20 | 1.52 ± 0.41 | 0.71 ± 0.11 |
| | 11 mm/s | 2.23 ± 0.63 | 3.09 ± 1.16 | 2.42 ± 0.81 | 2.34 ± 0.70 | 2.30 ± 0.37 | 1.02 ± 0.11 |
| 500 W | 9 mm/s | 2.22 ± 0.24 | 3.07 ± 0.32 | 2.63 ± 0.32 | 2.65 ± 0.33 | 1.89 ± 0.15 | 0.84 ± 0.08 |
| | 10 mm/s | 1.90 ± 0.27 | 2.71 ± 0.34 | 2.18 ± 0.45 | 2.17 ± 0.53 | 1.75 ± 0.28 | 0.69 ± 0.14 |
| | 11 mm/s | 2.89 ± 0.61 | 3.90 ± 0.94 | 3.30 ± 0.80 | 3.39 ± 0.79 | 2.73 ± 0.37 | 1.12 ± 0.12 |
| 550 W | 9 mm/s | 2.95 ± 0.85 | 4.04 ± 1.11 | 3.31 ± 0.89 | 3.31 ± 0.87 | 2.68 ± 0.94 | 1.42 ± 0.43 |
| | 10 mm/s | 2.32 ± 0.71 | 3.39 ± 1.02 | 2.78 ± 0.94 | 2.78 ± 0.95 | 1.88 ± 0.45 | 0.77 ± 0.20 |
| | 11 mm/s | 2.97 ± 0.63 | 3.82 ± 0.83 | 3.30 ± 0.82 | 3.33 ± 0.82 | 3.07 ± 0.41 | 1.36 ± 0.26 |

*Future directions.* Several remarks suggest further research directions. Our qualitative analysis (Figs. 10–12) reveals that prediction errors vary by material due to different thermophysical properties. We claim that this stems from the static collocation sampling strategy used in this study, which is independent of material properties. However, to the best of the authors' knowledge, adaptive sampling conditioned on the process condition has not been addressed in the metal AM literature. This, in turn, presents a promising future research direction: developing adaptive collocation sampling methods to enable accurate temperature field predictions across diverse process conditions.

Alternatively, integrating Fourier Neural Operators (FNO) could attenuate high residuals near the laser-irradiated zone by capturing high-frequency thermal features. By transforming the current spatiotemporal domain into the frequency domain, this may help capture the accurate temperature field near the high-frequency regions and handle generalization in a better way compared to the framework presented in this study.

In addition, extending the framework to incorporate process parameters alongside material properties as inputs represents an important direction. Although we validated our framework's robustness under different process parameter settings, retraining is currently required when process parameters change because only material properties are included as inputs. Since process parameters are significant factors in the monitoring and control of metal AM, the authors believe that a decoupled architecture, combined with the proposed physics-guided output scaling, will help demonstrate strong performance and enhance the flexibility that metal AM promises.

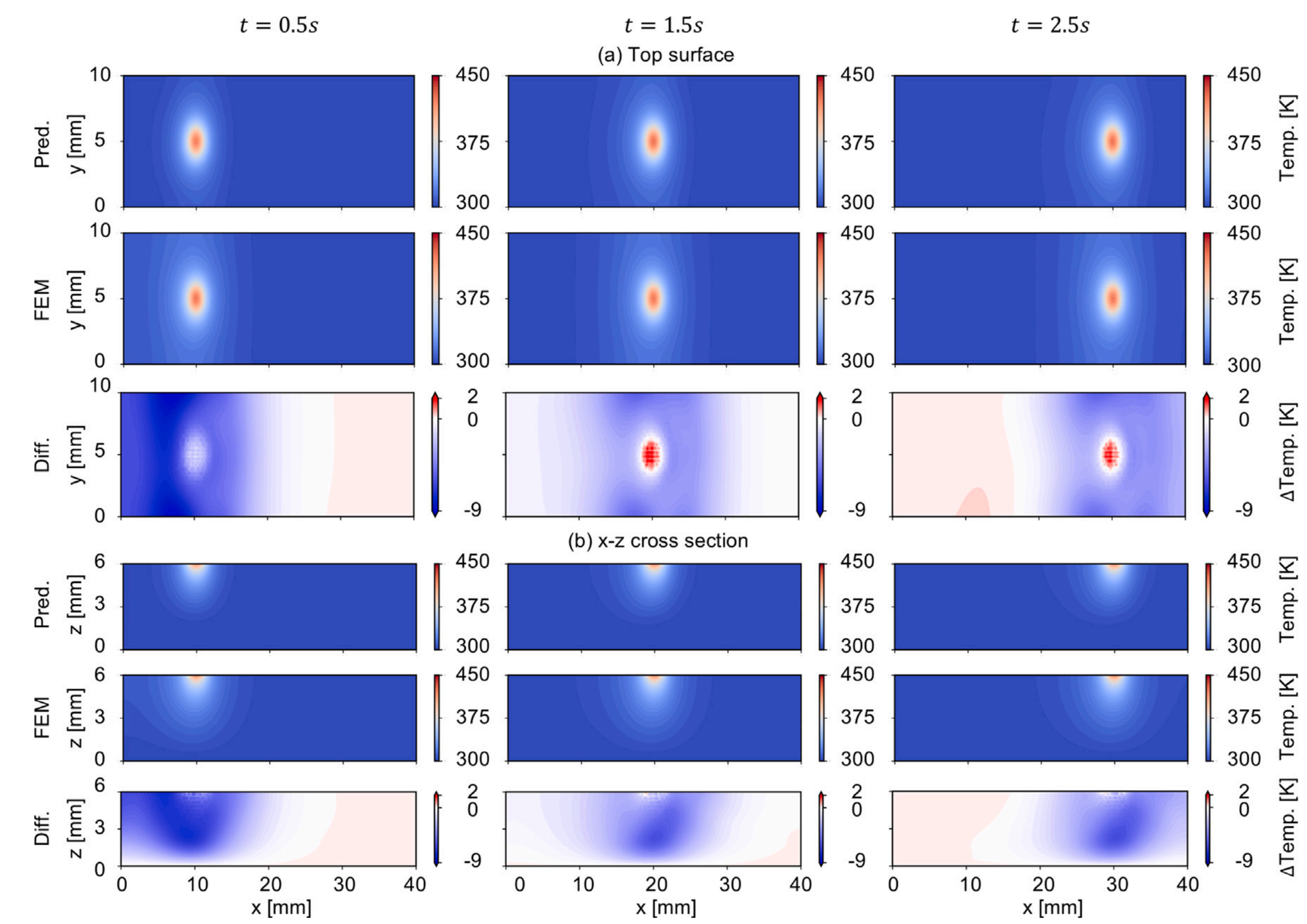

**Fig. 13.** Qualitative comparison of temperature fields at $t = 0.5, 1.5, 2.5$ s predicted by the proposed framework and FEM simulations for Copper. (a) Top-surface temperature fields predicted by the proposed framework, the corresponding FEM reference solutions, and the discrepancy $\Delta T = \hat{T}_{\text{phys}}(\Theta) - T_{\text{FEM}}$. (b) the corresponding $x$–$z$ cross-sections following the same ordering.

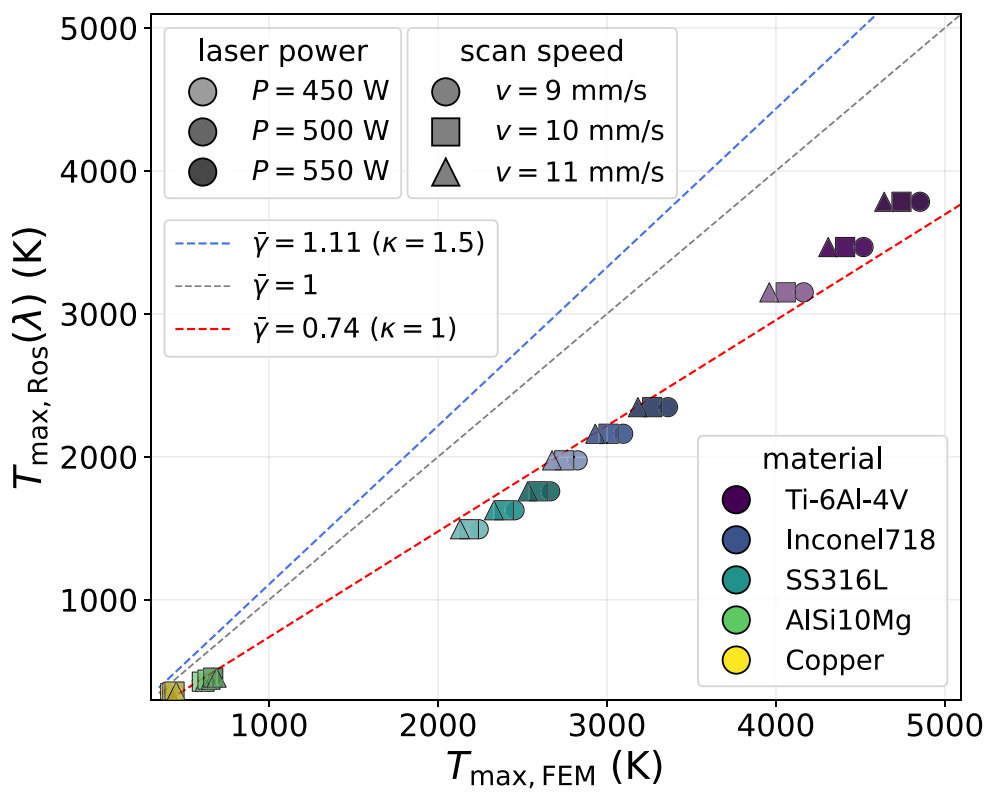

**Fig. 14.** Peak temperatures predicted by the Rosenthal solution vs. FEM across all process conditions and materials, where each point represents one (material, process parameter) combination. Without correction ($\kappa = 1$), the Rosenthal solution systematically underpredicts FEM peak temperatures by ∼26% ($\bar{\gamma} = 0.74$); scaling by $\kappa = 1.5$ raises this to $\bar{\gamma} = 1.11$, safely bounding the physically admissible range.

*Limitations.* Nevertheless, several limitations remain unresolved in this study and require further investigation. First, the current framework assumes temperature-independent, constant material properties, which may require additional calibration when applied in practice. For instance, when predicting temperature-dependent physical variables (e.g., residual stress), or when incorporating complex multiphysics such as thermo-fluid dynamics and phase transitions in metal AM, the assumption of constant material properties may slightly conflict with the physics. Although adopting temperature-independent properties is a common simplification in the existing literature, it becomes an inevitable assumption in our specific task to achieve temperature field predictions across unseen materials.

Furthermore, the experimental setup is based on a simplified numerical benchmark that is frequently used in the literature to describe the phenomena in metal AM processes. Although this setup serves as an effective surrogate model to characterize how temperature fields behave across different materials, the aforementioned limitations can introduce slight deviations from actual scenarios, requiring minor calibrations to align with reality. To better reflect real-world metal AM processes, external validation using physical experiments would be highly beneficial.

Lastly, this study assumes that only a single material is utilized during the process, which may limit its perceived direct impact on certain industrial practices. However, to the best of the authors' knowledge, thermal modeling for multi-material AM has not yet been addressed in the literature. Therefore, extending this framework to multi-material configurations represents a promising direction for future research.

## 5. Ablation study

In this section, we carry out ablation studies to validate the effectiveness of each component in the proposed framework.

### 5.1. Physics-guided output scaling

Table 7 presents an ablation study on different output-scaling strategies for material-agnostic temperature prediction. Specifically, directly using the raw network output $\hat{T}_\Theta$ leads to severely degraded performance with extremely large variance, indicating unstable training and physically implausible predictions. This failure arises because the network is not constrained to satisfy the fundamental thermal condition in metal AM that temperatures remain greater than or equal to the ambient temperature $T_\infty$.

**Table 7**
Ablation study on physics-guided output scaling strategies for material-agnostic temperature prediction. We report $L_2$ error [%] averaged over five random seeds as mean ± standard deviation. Lower values indicate better performance.

| $\hat{T}_{\text{phys}}(\Theta)$ | $T_{\max}$ | Ti–6Al–4V (4430, 560, 6.7) | Inconel 718 (8220, 435, 11.4) | SS 316L (8000, 500, 16) |
|---|---|---|---|---|
| $\hat{T}_\Theta$ | – | $77.81 \pm 31.91$ | $74.03 \pm 37.29$ | $72.39 \pm 39.62$ |
| $T_\infty + \text{Softplus}(\hat{T}_\Theta)$ [Eq. (9)] | – | $38.59 \pm 0.00$ | $27.90 \pm 0.00$ | $23.05 \pm 0.00$ |
| $T_\infty + T_{\max} \cdot \text{Softplus}(\hat{T}_\Theta)$ [Eq. (8)] | 600 | $13.56 \pm 10.64$ | $10.53 \pm 7.86$ | $9.30 \pm 6.52$ |
| | 3000 | $5.66 \pm 1.45$ | $4.43 \pm 1.35$ | $4.10 \pm 1.23$ |
| | 15 000 | $6.67 \pm 2.29$ | $5.51 \pm 1.36$ | $5.09 \pm 1.02$ |
| $T_\infty + T_{\max}(\lambda;\Theta) \cdot \text{Softplus}(\hat{T}_\Theta)$ | $g_\theta(\lambda_{\text{norm}})$ | $38.50 \pm 0.09$ | $27.85 \pm 0.05$ | $23.01 \pm 0.04$ |
| $T_\infty + \kappa \cdot T_{\max}(\lambda) \cdot \text{Softplus}(\hat{T}_\Theta)$ [Eq. (12)] | **Proposed** | $\mathbf{2.71} \pm 0.34$ | $\mathbf{2.18} \pm 0.45$ | $\mathbf{2.17} \pm 0.53$ |

Introducing a simple positivity constraint using $T_\infty$ and a softplus offset, $T_\infty + \text{Softplus}(\hat{T}_\Theta)$, improves numerical stability but consistently converges to trivial solutions. Although this formulation enforces the admissible thermal constraint $T \geq T_\infty$, it fails to capture the complex thermal behavior induced by laser heating at the top boundary condition. As a result, the network collapses to a non-physical solution that satisfies the constraint but ignores the governing thermal dynamics. This behavior arises because the network must solely represent temperature fields spanning a very wide physical range — from the ambient temperature (= 300 K) to peak temperatures of several thousand kelvins — without any explicit scaling. Learning such a large dynamic range poses a severe optimization challenge, leading to unstable gradient magnitudes and significant training instability.

Extending this approach as in prior works [26,34] by adopting a manually specified constant upper bound $T_{\max}$ provides only limited improvements and remains highly sensitive to hyperparameter selection. Moderate values (e.g., $T_{\max} = 3000$) yield the best performance, whereas underestimation (e.g., $T_{\max} = 600$) leads to significant instability because the imposed scale fails to cover the high peak temperatures generated by laser heating. In this case, similar to the (9), the network must compensate by producing extremely large values of $\text{Softplus}(\hat{T}_\Theta)$ to match the true temperature scale, which results in poorly conditioned gradients and unstable optimization. Conversely, large overestimations (e.g., $T_{\max} = 15000$) restrict the effective range of $\text{Softplus}(\hat{T}_\Theta)$ to a relatively narrow interval. Although this limits the expressiveness of the scaling and leads to slightly degraded accuracy, it produces better numerical stability than underestimation because the network operates within a more moderate activation range during optimization.

Although a previous study reported that overestimating $T_{\max}$ does not significantly degrade performance in non-parametric PINNs trained for a single material [44], this assumption does not hold in cross-material generalization scenarios. In material-agnostic inference, where the model must generalize across unseen material properties, manually selecting a universal temperature scale becomes unreliable and leads to significant performance degradation. Moreover, determining an appropriate $T_{\max}$ for all material configurations is impractical without, for instance, prior access to ground-truth simulations.

We further investigate an alternative by introducing a small auxiliary network $g_\theta(\lambda_{\text{norm}})$ to predict $T_{\max}$ directly from material properties. However, this scaling strategy introduces additional optimization burden to the network, where multi-objectives already exist, and task interference due to its stochastic nature. As a result, the model frequently converges to trivial solutions similar to those observed in the unconstrained scaling case of Eq. (9).

In contrast, the proposed physics-guided output scaling deterministically constructs a material-conditioned temperature scale grounded in Rosenthal's analytical solution without introducing additional learnable components. This deterministic, physics-informed scaling stabilizes optimization, reduces network complexity, and allows the model to focus on learning a material-agnostic representation of thermal behavior. Consequently, the proposed approach achieves the best performance with consistently minute variance across all materials.

**Table 8**
Sensitivity analysis of the proposed physics-guided output scaling strategy w.r.t. the scaling factor $\kappa$. We report $L_2$ error [%] averaged over five random seeds as mean ± standard deviation. Lower values indicate better performance.

| Material $(\rho, C_p, k)$ | Ti–6Al–4V (4430, 560, 6.7) | Inconel 718 (8220, 435, 11.4) | SS 316L (8000, 500, 16) |
|---|---|---|---|
| *Baselines* | | | |
| N-PINN | $6.19 \pm 1.00$ | $6.09 \pm 0.83$ | $5.94 \pm 0.63$ |
| P-PINN | $7.59 \pm 3.86$ | $6.12 \pm 3.22$ | $5.84 \pm 2.59$ |
| *Proposed method* | | | |
| $\kappa = 1.0$ | $9.67 \pm 13.18$ | $7.39 \pm 9.61$ | $6.62 \pm 7.89$ |
| $\kappa = 1.25$ | $6.56 \pm 3.28$ | $5.38 \pm 2.51$ | $5.23 \pm 2.13$ |
| $\boldsymbol{\kappa = 1.5}$ | $\mathbf{2.71 \pm 0.34}$ | $\mathbf{2.18 \pm 0.45}$ | $\mathbf{2.17 \pm 0.53}$ |
| $\kappa = 1.75$ | $4.51 \pm 1.53$ | $3.75 \pm 1.28$ | $3.71 \pm 1.39$ |
| $\kappa = 2.0$ | $3.83 \pm 1.42$ | $3.21 \pm 1.27$ | $3.16 \pm 1.31$ |

Finally, we conduct a sensitivity analysis on the scaling factor $\kappa \geq 1$ used in the proposed physics-guided output scaling. Recall that $\kappa$ is introduced to compensate for the inherent underestimation of the peak temperature obtained from Rosenthal's analytical solution when applied to metal AM processes due to the simplifying assumptions underlying the analytical model. Throughout this study, the scaling factor is set to $\kappa = 1.5$. Table 8 summarizes the sensitivity results.

When $\kappa$ is set to smaller values (e.g., $\kappa = 1.0$ or 1.25), the resulting upper bound $T_{\max}$ underestimates the true peak temperature. As a consequence, the network must compensate by producing excessively large values of the normalized output to match the actual temperature scale, which increases the optimization burden and leads to training instability. This behavior is reflected in Table 8, where $\kappa = 1.0$ results in both large prediction errors and extremely high variance. Although the performance improves somewhat at $\kappa = 1.25$, the model still exhibits noticeable instability.

In contrast, setting $\kappa = 1.5$ provides the closest upper bound that better covers the true temperature range. This reduces the burden on the network to represent extreme temperature magnitudes, resulting in the most stable training behavior and the best predictive accuracy with minimal variance across all materials. As $\kappa$ increases further (e.g., $\kappa = 1.75$ or 2.0), the upper bound becomes increasingly overestimated. Although this introduces a mismatch between the scaling and the true temperature range, the resulting degradation in accuracy remains relatively mild compared with the severe instability observed under underestimation. As shown in both Table 8 and Fig. 15, the proposed method consistently outperforms the baseline models even for these larger values of $\kappa$.

To sum up, these results highlight the importance of approximating the peak temperature when performing material-agnostic thermal modeling. Requiring the network to directly represent the full temperature

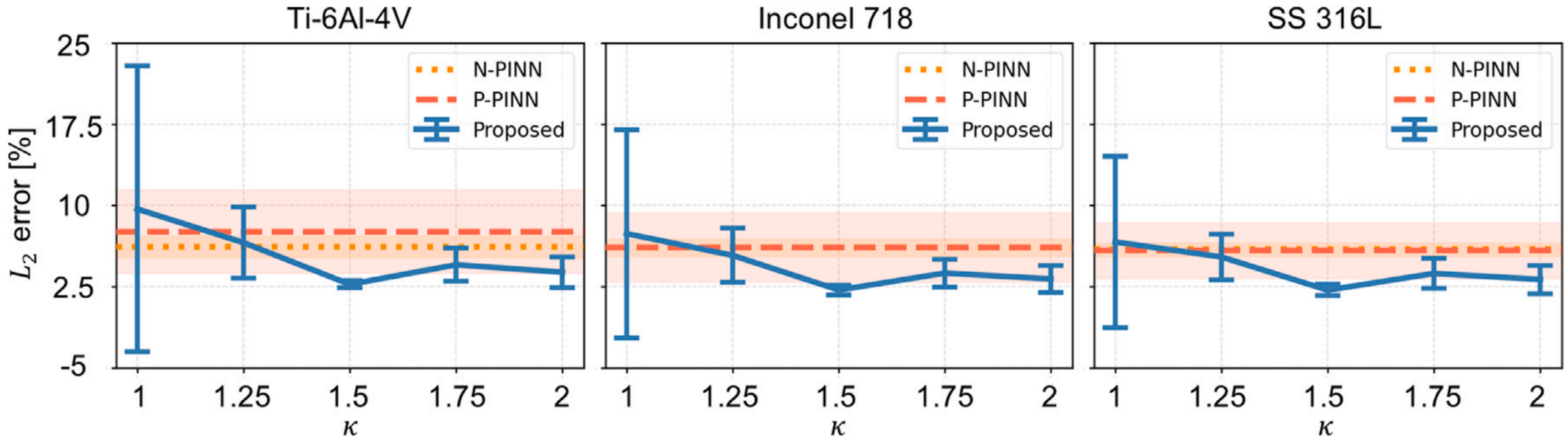

**Fig. 15.** Sensitivity analysis of the proposed physics-guided output scaling strategy w.r.t. the scaling factor $\kappa$. Solid lines represent the mean value averaged over five random seeds and the shaded regions indicate ± one standard deviation.

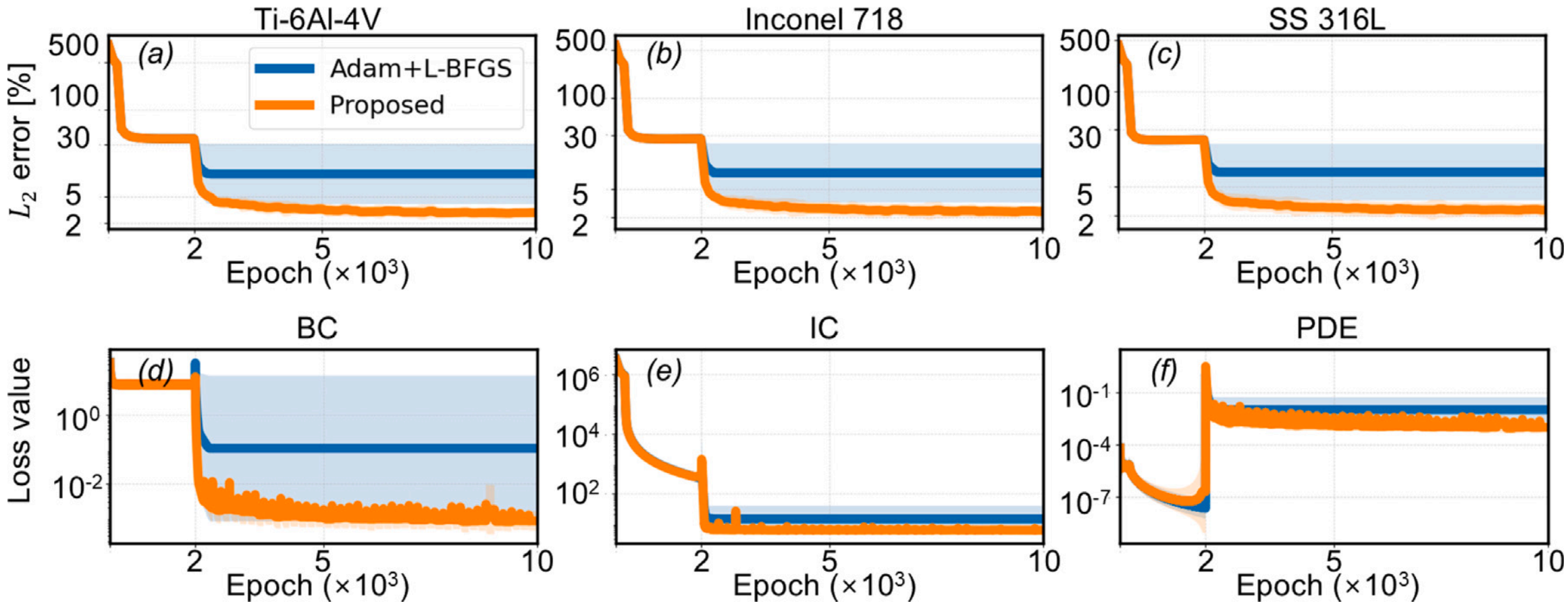

**Fig. 16.** Comparison of training dynamics between the Adam+L-BFGS and the proposed hybrid training strategy. (a)–(c) Evolution of the $L_2$ error [%] during training for Ti–6Al–4V, Inconel 718, and SS 316L. (d)–(f) Evolution of the loss components in (2): boundary condition loss (4), initial condition loss (5), and PDE residual loss (3). Solid lines represent the mean over five random seeds, and shaded regions indicate ± one standard deviation.

**Table 9**
Ablation study on the synergistic effect of the proposed loss optimization strategy for the kinds of PINNs in metal AM. We report $L_2$ error [%] averaged over five random seeds as mean ± standard deviation. Lower values indicate better performance.

| Method | Loss optimization | Ti–6Al–4V (4430, 560, 6.7) | Inconel 718 (8220, 435, 11.4) | SS 316L (8000, 500, 16) |
|---|---|---|---|---|
| **Proposed** | Adam + L-BFGS | 18.21 ± 18.26 | 13.49 ± 12.93 | 11.54 ± 10.33 |
| | **Proposed** | **2.71** ± 0.34 | **2.18** ± 0.45 | **2.17** ± 0.53 |

**Table 10**
Ablation study on the synergistic effect of the proposed loss optimization strategy for the kinds of PINNs in metal AM. We report $L_2$ error [%] averaged over five random seeds as mean ± standard deviation. Lower values indicate better performance.

| Method | Loss optimization | Ti–6Al–4V (4430, 560, 6.7) | Inconel 718 (8220, 435, 11.4) | SS 316L (8000, 500, 16) |
|---|---|---|---|---|
| N-PINN | Adam | 6.19 ± 1.00 | **6.09** ± 0.83 | 5.94 ± 0.63 |
| | **Proposed** | **5.60** ± 2.47 | 6.56 ± 2.14 | **4.58** ± 2.00 |
| P-PINN | Adam | 16.66 ± 2.01 | 13.14 ± 1.17 | 11.95 ± 0.92 |
| | **Proposed** | **7.59** ± 3.86 | **6.12** ± 3.22 | **5.84** ± 2.59 |
| **Proposed** | Adam | 26.49 ± 8.05 | 19.01 ± 5.78 | 15.73 ± 4.64 |
| | **Proposed** | **2.71** ± 0.34 | **2.18** ± 0.45 | **2.17** ± 0.53 |

range — from $T_\infty$ to extreme peak temperatures $T_{max}$ — introduces a significant optimization burden and often leads to training instability. In contrast, incorporating a physics-guided approximation of the peak temperature through output scaling substantially resolves this burden, enabling stable training and convergence to accurate, non-trivial solutions across any unseen materials.

### 5.2. Hybrid optimization

Table 9 evaluates the impact of the proposed loss optimization strategy against the conventional Adam+L-BFGS approach. In both settings, Adam is applied for the first 2000 epochs, after which different L-BFGS variants are used. While conventional L-BFGS relies on full-batch curvature estimation, our method adopts stochastic mini-batch sampling of collocation points with resampling triggered by changes in curvature information. The results show that the proposed strategy consistently achieves lower error and variance across all three materials.

Fig. 16 illustrates the performance difference. After the Adam phase until 2000 epochs, the conventional method shows little improvement in both loss and $L_2$ error, likely due to the static curvature information obtained from full-batch updates. In contrast, the proposed method employs stochastic mini-batch sampling of collocation points, introducing mild stochasticity that facilitates optimization. Moreover, collocation points are resampled when curvature information remains unchanged across epochs, enabling continued curvature updates and more effective loss minimization, which leads to a substantial reduction in $L_2$ error.

Table 10 evaluates the impact of the proposed loss optimization strategy across several PINN variants commonly used in metal AM. Notably, the proposed strategy is trained for only 10,000 epochs, whereas the Adam baseline uses 50,000 epochs. While the proposed framework along with this strategy achieves the best performance, the results on the baselines also indicate that the benefits of the optimization scheme extend beyond the proposed framework to other PINN formulations.

Specifically, when applied to N-PINN and P-PINN baselines, the proposed strategy consistently yields lower $L_2$ errors than solely using

Adam optimizer in most cases. This improvement across different backbones demonstrates the broad applicability of the method, especially for parametric formulations where the loss landscape is often stiff and prone to instability. These findings suggest that the proposed strategy mitigates common training challenges in PINN-based thermal modeling for metal AM by significantly reducing extremely long training horizons, benefiting future PINN research regardless of the underlying network architecture.

## 6. Conclusion

This study addresses the challenge of material-agnostic thermal modeling for metal AM without requiring labeled data, retraining, or pre-training. The proposed framework builds on a parametric PINN formulation that incorporates three key components: (i) a decoupled parametric PINN architecture with FiLM-based conditioning for improved representation learning and physics-inductive bias, (ii) physics-guided output scaling derived from Rosenthal's analytical solution, and (iii) a hybrid optimization strategy designed to alleviate the training instability and long training horizons typical of conventional PINNs.

Extensive experiments demonstrate that the proposed framework achieves high accuracy and efficiency across diverse materials, including challenging OOD cases, enabling material-agnostic temperature prediction. We further show that parametric PINNs outperform non-parametric PINNs by avoiding material-specific overfitting and providing more accurate thermal modeling. Qualitative analyses additionally reveal how thermophysical material properties influence PINN behavior and generalization.

Finally, rigorous ablation studies and sensitivity analyses identify training instability in parametric PINNs as arising from the increased representational burden of the network, which is addressed by the proposed physics-guided output scaling. We additionally demonstrate that the proposed hybrid optimization strategy effectively mitigates the extremely long training horizons typically observed in PINNs. Last but not least, the modular design of the framework — including the network architecture, output scaling, and optimization strategy — provides a scalable foundation that can be adapted to various scenarios, contributing to the advancement of PINN-based thermal modeling in metal AM.

Future work may explore more advanced adaptive sampling strategies, including material-aware sampling as discussed earlier, to better account for the influence of thermophysical material properties in parametric PINNs. Furthermore, extending the proposed framework to incorporate additional process parameters, such as laser power, scanning speed, or laser paths, within the proposed decoupled architecture combined with physics-guided output scaling could further improve generalization to unseen process conditions.

Nonetheless, several limitations should be acknowledged. The current framework assumes temperature-independent material properties, which may require calibration when applied to real-world scenarios. In addition, the present study focuses solely on thermal modeling; incorporating more complex physics such as thermo-fluid dynamics and phase transitions could provide deeper insights, particularly in melt pool regions where extreme thermal gradients are induced by laser irradiation. Furthermore, the present evaluation relies solely on FEM-generated reference fields; experimental thermal validation remains an important limitation and is identified as a prioritized direction for future work. Lastly, this work considers only single-material AM. Our future research will investigate the applicability of the proposed framework to multi-material AM systems.

## CRediT authorship contribution statement

**Hyeonsu Lee:** Writing – original draft, Visualization, Validation, Software, Methodology, Investigation, Formal analysis, Data curation, Writing – review & editing. **Jihoon Jeong:** Writing – review & editing, Supervision, Methodology, Investigation, Funding acquisition, Conceptualization.

## Declaration of competing interest

The authors declare that they have no known competing financial interests or personal relationships that could have appeared to influence the work reported in this paper.

## Acknowledgment

This work was supported by the National Research Foundation of Korea (NRF) grant funded by the Korea government (MSIT) (RS-2025-02216260) and a grant of the Basic Research Program funded by the Korea Institute of Machinery and Materials (grant number: NK254A, Project Title: Development of Core Technologies for Advanced Chiplet Packaging Equipment).

## Appendix

### A.1. Nomenclature

See Table 11.

**Table 11**
List of symbols used in this paper.

| Symbol | Description | Unit |
|---|---|---|
| $\Omega$ | Spatial domain | $mm^3$ |
| $\partial\Omega$ | Boundary of spatial domain | – |
| $\mathcal{D}$ | Spatiotemporal domain $\Omega \times [0, t_{\text{end}}]$ | – |
| $\mathcal{M}$ | Material property space | – |
| $\lambda$ | Material property vector | – |
| $\lambda_m$ | $m$th material property component | – |
| $T(\mathbf{x}, t, \lambda)$ | Temperature field | K |
| $\hat{T}_\Theta$ | Predicted raw temperature field | K |
| $\hat{T}_{\text{phys}}$ | Predicted physically admissible temperature field | K |
| $\Theta$ | Set of learnable network parameters | – |
| $\mathcal{F}_\Theta$ | Parametric PINN mapping function | – |
| $g_{\theta_{\mathbf{x},t}}$ | Spatiotemporal feature extraction network | – |
| $g_{\theta_\lambda}$ | Material property encoder | – |
| $g_{\theta_c}$ | Fusion network | – |
| $\mathcal{N}(\cdot)$ | PDE residual operator | – |
| $\mathcal{B}(\cdot)$ | Boundary condition operator | – |
| $\mathcal{I}(\cdot)$ | Initial condition operator | – |
| $\mathcal{L}$ | Total PINN loss function | – |
| $\mathcal{L}_{\text{pde}}$ | PDE residual loss | – |
| $\mathcal{L}_{\text{bc}}$ | Boundary condition loss | – |
| $\mathcal{L}_{\text{ic}}$ | Initial condition loss | – |
| $w_{\text{pde}}, w_{\text{bc}}, w_{\text{ic}}$ | Loss weights | – |
| $N_{\text{pde}}$ | Number of PDE collocation points | – |
| $N_{\text{bc}}$ | Number of boundary collocation points | – |
| $N_{\text{ic}}$ | Number of initial condition collocation points | – |
| $\rho$ | Density | kg $m^{-3}$ |
| $C_p$ | Specific heat capacity | J $kg^{-1}$ $K^{-1}$ |
| $k$ | Thermal conductivity | W $m^{-1}$ $K^{-1}$ |
| $\alpha$ | Thermal diffusivity ($\alpha = k/(\rho C_p)$) | $m^2$ $s^{-1}$ |
| $T_0$ | Initial temperature | K |
| $T_\infty$ | Ambient temperature | K |
| $T_{\max}$ | Estimated maximum temperature | K |
| $\kappa$ | Scaling factor for $T_{\max}$ | – |
| $\eta$ | Laser absorptivity | – |
| $P$ | Laser power | W |
| $r_{\text{laser}}$ | Laser beam radius | m |
| $v$ | Laser scanning speed | m $s^{-1}$ |
| $\mathbf{x}_{\text{laser}}(t)$ | Laser center position | m |
| $q_{\text{laser}}$ | Laser heat flux | W $m^{-2}$ |
| $q_{\text{conv}}$ | Convective heat flux | W $m^{-2}$ |
| $q_{\text{rad}}$ | Radiative heat flux | W $m^{-2}$ |
| $h$ | Convection heat transfer coefficient | W $m^{-2}$ $K^{-1}$ |
| $e$ | Surface emissivity | – |
| $\sigma$ | Stefan–Boltzmann constant | W $m^{-2}$ $K^{-4}$ |
| $\mathbf{n}$ | Outward unit normal vector | – |

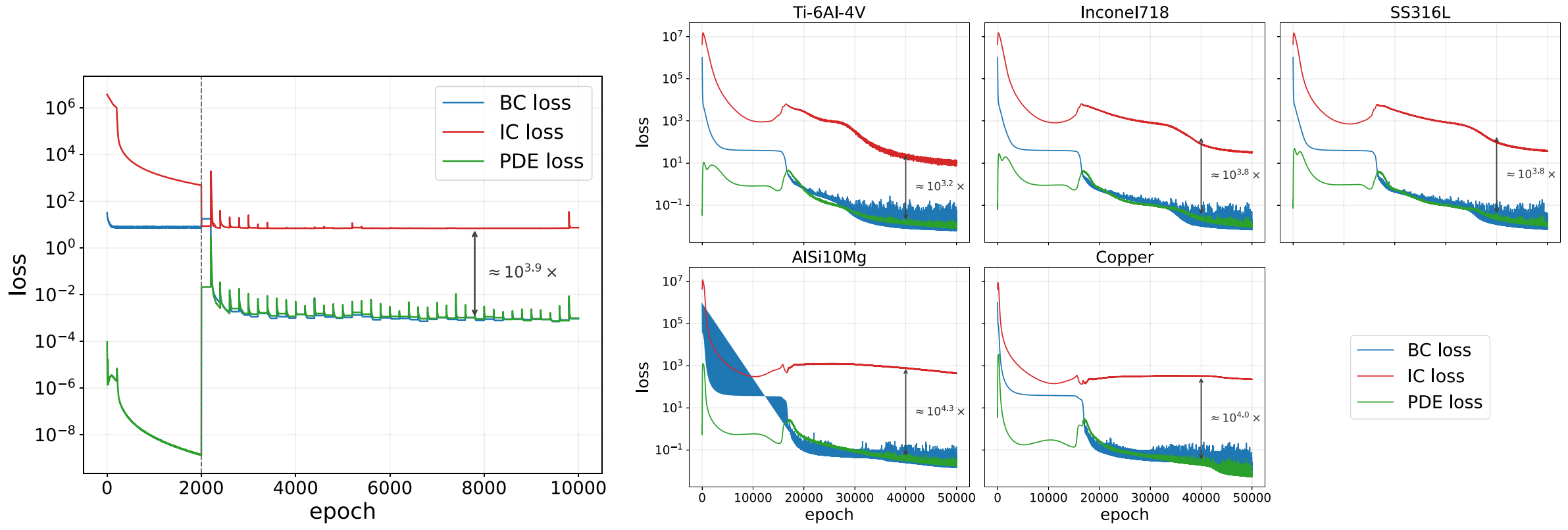

(a) Loss trend in the proposed framework.

(b) Loss trends in the N-PINN baseline across five materials.

**Fig. 17.** Optimization dynamics during PINN training for (a) the proposed framework and (b) the N-PINN baseline. Both models exhibit a shared trajectory: initially drifting toward a non-physical trivial solution (where $\mathcal{L}_{ic}$ and $\mathcal{L}_{pde}$ dominate $\mathcal{L}_{bc}$) before successfully transitioning and converging toward a physically consistent, balanced loss scaling regime.

### *A.2. Training algorithm*

**Algorithm 1** Training the Proposed Framework for Material-Agnostic Temperature Prediction

**Input:** Spatiotemporal domain $\mathcal{D}$, material property space $\mathcal{M}$
**Parameters:** Scale factor $\kappa$; collocations $N_{bc}, N_{pde}, N_{ic}$; loss weights $w_{pde}, w_{bc}, w_{ic}$; learning rates $lr_{Adam}, lr_{L-BFGS}$; epochs $J_{Adam}, J_{L-BFGS}$; batch size $|B|$; process parameters $P, \eta, r_{laser}, v, \mathbf{x}_0, T_0, T_\infty, h, e$
**Output:** Trained model $\mathcal{F}_\Theta$ for temperature field inference over any material $\lambda$
Initialize $\Theta^{(0)} = \{\theta^{(0)}_{\mathbf{x},t}, \theta^{(0)}_{\lambda}, \theta^{(0)}_{c}\}$; sample collocation points from $\mathcal{D} \times \mathcal{M}$
// **Phase 1: Adam**
**for** $j = 1$ to $J_{Adam}$ **do**
  Sample $B \subset \mathcal{D} \times \mathcal{M}$
  Compute $\hat{T}_{phys}(\Theta)$ via Eq. (12) and $\mathcal{L}_\Theta$ via Eq. (2)
  Update $\Theta^{(j)}$ via Eq. (13)
**end for**
// **Phase 2: Mini-batch L-BFGS**
**for** $j = 1$ to $J_{L-BFGS}$ **do**
  Sample $B \subset \mathcal{D} \times \mathcal{M}$
  Compute $\hat{T}_{phys}(\Theta)$ via Eq. (12) and $\mathcal{L}_\Theta$ via Eq. (2)
  Update $\Theta^{(j)}$ via Eq. (14)
  **if** $\mathbf{H}_j^{-1} \approx \mathbf{H}_{j-1}^{-1}$ **then**
    Resample $\mathcal{B}$ from $\mathcal{D} \times \mathcal{M}$
  **end if**
**end for**
**return** $\Theta^{(J_{Adam}+J_{L-BFGS})}$

### *A.3. Governing equations and boundary conditions*

The transient heat transfer in the bare plate metal AM process is governed by the conservation of energy coupled with Fourier's law of heat conduction. The temperature field $T(\mathbf{x}, t; \lambda)$ satisfies

$$\text{Governing PDE:} \quad \rho C_p \frac{\partial T}{\partial t} - \nabla \cdot \left(k \nabla T\right) = 0 \qquad \mathbf{x} \in \Omega,\ t > 0$$

$$\text{Initial condition:} \quad T(\mathbf{x}, 0; \lambda) = T_0 \qquad \mathbf{x} \in \Omega$$

The boundary $\partial\Omega$ is decomposed into the top surface $\Gamma_{top}$, the side surfaces $\Gamma_{side}$, and the bottom surface $\Gamma_{bottom}$. The corresponding boundary conditions are specified as follows.

$$\text{Top surface:} \quad -k\nabla T \cdot \mathbf{n} = q_{laser} + q_{conv} + q_{rad} \qquad \mathbf{x} \in \Gamma_{top},\ t > 0$$

$$\text{Side surfaces:} \quad -k\nabla T \cdot \mathbf{n} = q_{conv} + q_{rad} \qquad \mathbf{x} \in \Gamma_{side},\ t > 0$$

$$\text{Bottom surface:} \quad T(\mathbf{x}, t; \lambda) = T_0 \qquad \mathbf{x} \in \Gamma_{bottom},\ t > 0$$

Here, the laser heat flux is imposed on the top surface. The applied laser heat flux is modeled using a Gaussian surface heat source,

$$q_{laser}(\mathbf{x}, t) = \frac{2\eta P}{\pi r_{laser}^2} \exp\left(-\frac{2\|\mathbf{x} - \mathbf{x}_{laser}(t)\|^2}{r_{laser}^2}\right)$$

where $\eta$ denotes the laser absorptivity, $P$ is the laser power, $r_{laser}$ is the laser beam radius, and $\mathbf{x}_{laser}(t)$ is the time-dependent laser center position. The laser center follows a prescribed scanning path given by $\mathbf{x}_{laser}(t) = \mathbf{x}_0 + \mathbf{v}\,t$. In this study, a one-dimensional scan along the $x$-axis on the top surface is considered, i.e., $\mathbf{v} = (v, 0, 0)$.

Convective and radiative heat fluxes are applied to all remaining surfaces except the substrate bottom, where a fixed-temperature Dirichlet boundary condition is specified. The convective and radiative heat fluxes are given by

$$q_{conv}(\mathbf{x}, t) = h\left(T(\mathbf{x}, t) - T_0\right)$$
$$q_{rad}(\mathbf{x}, t) = \sigma e\left(T(\mathbf{x}, t)^4 - T_0^4\right)$$

where $h$ is the convection heat transfer coefficient, $\sigma$ is the Stefan–Boltzmann constant, and $e$ is the surface emissivity.

### *A.4. Implementation details*

All experiments were conducted on a single NVIDIA RTX 5090 GPU using Python 3.10 and PyTorch. The mean training and inference times of the proposed framework were approximately 6800 s and 9.2 ms, respectively.

The total number of collocation points sampled from the entire input domain during training is 555,338, consisting of $N_{BC} = 137,555$, $N_{IC} = 909$, and $N_{PDE} = 416,874$. The boundary points are distributed as follows: $N_{+x} = 4697$, $N_{+y} = 17,507$, $N_{-x} = 4697$, $N_{-y} = 17,507$, $N_{\Gamma_{top}} = 65,636$, and $N_{\Gamma_{bottom}} = 27,511$.

During training, the mini-batch sizes $|B|$ for the Adam phase are $|B|_{BC} = 12,000$, $|B|_{IC} = 6000$, and $|B|_{PDE} = 20,000$. For the L-BFGS phase, the mini-batch sizes are $|B|_{BC} = 8000$, $|B|_{IC} = 4000$, and $|B|_{PDE} = 12,000$.

As aforementioned, the material property vector $\lambda$ is sampled uniformly at random from the admissible space $\mathcal{M}$ for each mini-batch during training, independently of the spatiotemporal collocation points.

All networks use the hyperbolic tangent (tanh) activation function. To prevent convergence to degenerate solutions where the boundary condition loss is minimized by collapsing the temperature field to the ambient value (i.e., trivial local optima), a small positive offset $\epsilon$ is added to the Softplus activation in Eq. (12): $\mathrm{Softplus}(\hat{T}_\Theta) + \epsilon, \quad \epsilon = 10^{-3}$

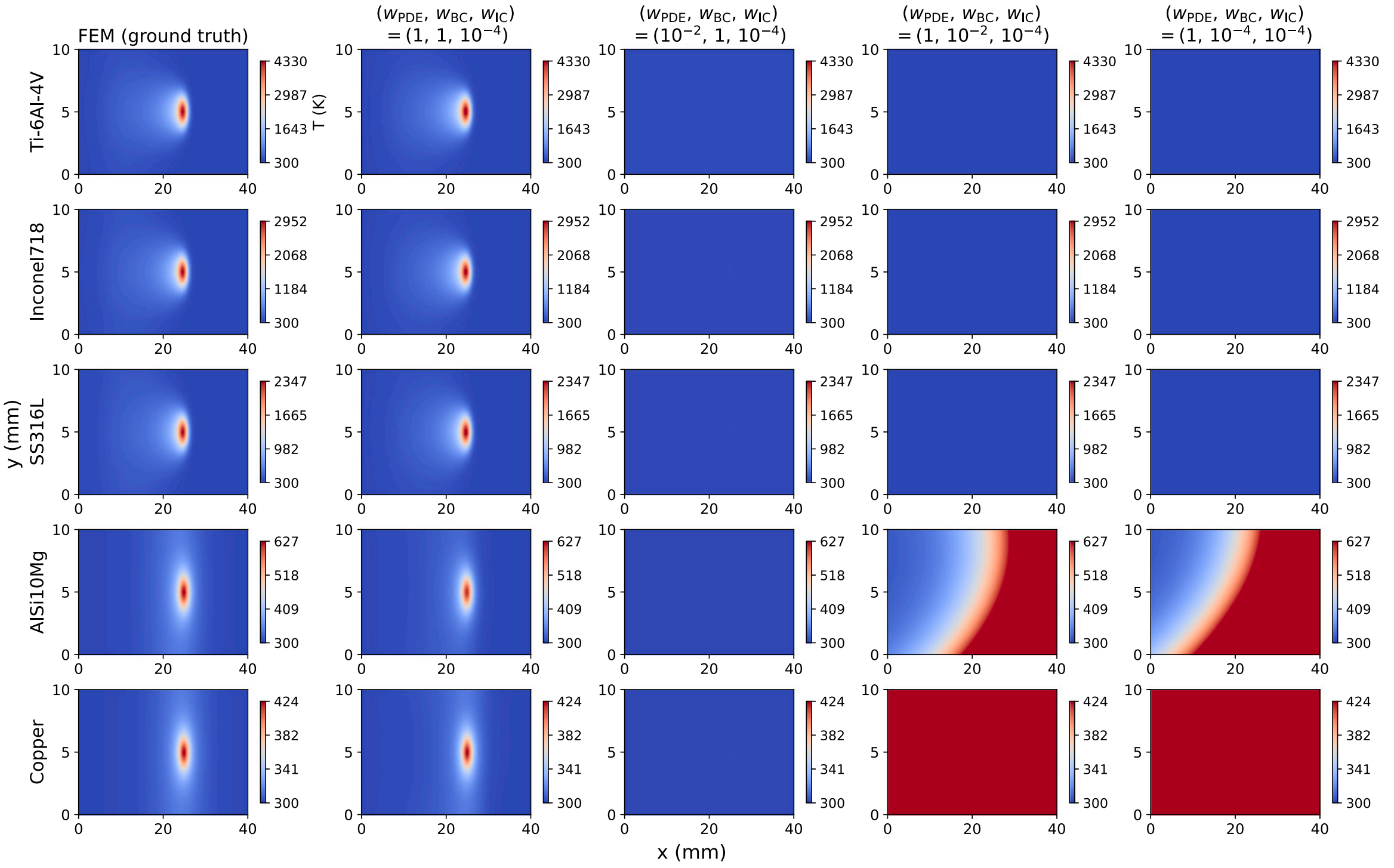

**Fig. A.18.** Top boundary temperature field at $t = 2.0$ s for the five evaluation materials (rows). Columns: FEM ground truth and PINN predictions for different loss weights $(w_{\text{pde}}, w_{\text{bc}}, w_{\text{ic}})$. All loss weight combinations other than the baseline $(w_{\text{pde}}, w_{\text{bc}}, w_{\text{ic}}) = (1, 1, 10^{-4})$ collapses the solution.

To prevent numerical overflow caused by the fourth-power radiation term, we implement a clipping mechanism to clip $\hat{T}_{\text{phys}}(\Theta)$ at implausibly high values (e.g., $10^6$ K), which ensures that the high-order derivatives do not destabilize the optimization, allowing for robust convergence.

Similarly to the spatiotemporal coordinates, material properties are normalized to the range $[-1, 1]$ using the known bounds $[\lambda_{\text{min}}, \lambda_{\text{max}}]$ before being fed into the material branch network to ensure stable training.

For the L-BFGS phase, the optimizer is configured with a learning rate of 1.0, a maximum of 50 quasi-Newton iterations per step, a history size of 50, and the strong Wolfe line-search condition.

### *A.5. Analysis of loss weights and optimization dynamics*

This subsection discusses the selection and impact of the loss weights that are configured as $(w_{\text{pde}}, w_{\text{bc}}, w_{\text{ic}}) = (1, 1, 10^{-4})$ in this study. Loss weights in PINNs are generally problem-specific, and we therefore employed the same weight combination from prior research (e.g., [26]). However, to the best of the authors' knowledge, a justification for the choice has often been omitted in the literature. To this end, we offer an in-depth analysis of the optimization dynamics to justify this configuration.

Through our empirical observations, we discovered that both the non-parametric baseline and the proposed framework follow a consistent pattern when successfully converging toward physically consistent, high-quality local optima. Specifically, when learning a physically valid solution, the PDE loss $\mathcal{L}_{\text{pde}}$ and BC loss $\mathcal{L}_{\text{bc}}$ maintain a similar scale, whereas the IC loss $\mathcal{L}_{\text{ic}}$ remains larger by a factor of approximately $10^4$ as shown in Fig. 17.

Conversely, when the model converges toward a physically invalid, trivial solution (i.e., a poor local optimum), a contrasting pattern emerges where only the PDE loss drops drastically. This phenomenon represents a well-known failure mode in PINNs, where the gradient of the PDE loss dominates those of the BC and IC losses, causing the optimizer to exclusively fit the residual equations while ignoring boundary and initial constraints [53]. We attribute this behavior to the competition between two primary optimization trajectories:

(1) The first trajectory leads to a physically meaningful solution, where the model successfully captures the steep thermal gradients induced by the laser heat source at the boundary, as well as the subsequent thermal diffusion. Along this path, the loss terms naturally scale at a ratio of approximately $(\mathcal{L}_{\text{pde}}, \mathcal{L}_{\text{bc}}, \mathcal{L}_{\text{ic}}) \approx (1, 1, 10^4)$. To prevent the IC loss from dominating the overall gradient descent and disrupting this balance, the scaling weights must be balanced accordingly, which justifies our choice of $w_{\text{ic}} = 10^{-4}$.

(2) The second trajectory occurs when the model collapses into a trivial, non-physical solution across the entire spatiotemporal domain. For instance, the model can easily minimize the PDE loss by predicting a uniform ambient temperature field throughout the domain. While this trivial solution satisfies the initial condition and fits materials with high thermal conductivity reasonably well, it largely violates the boundary conditions (particularly at the top surface irradiated by the laser). In this failure mode, the loss scales become significantly imbalanced; the BC loss fails to decrease, while the PDE and IC losses drop to extremely low values.

**Table 12**
Effect of loss weights on mean relative $L_2$ error (%) across the five evaluation materials over five seeds for each $(w_{\text{bc}}, w_{\text{pde}})$ combination with $w_{\text{ic}} = 10^{-4}$ fixed. Only the baseline $(w_{\text{pde}}, w_{\text{bc}}, w_{\text{ic}}) = (1, 1, 10^{-4})$ converges to a physically consistent, non-trivial solution.

| $(w_{\text{pde}}, w_{\text{bc}}, w_{\text{ic}})$ | $L_2$ error (%) |
|---|---|
| $(10^{-4}, 10^{-4}, 10^{-4})$ | $71.6 \pm 30.0$ |
| $(10^{-2}, 10^{-4}, 10^{-4})$ | $71.5 \pm 30.0$ |
| $(1, 10^{-4}, 10^{-4})$ | $71.4 \pm 30.0$ |
| $(10^{-4}, 10^{-2}, 10^{-4})$ | $51.2 \pm 28.0$ |
| $(10^{-2}, 10^{-2}, 10^{-4})$ | $51.2 \pm 28.0$ |
| $(1, 10^{-2}, 10^{-4})$ | $51.2 \pm 28.0$ |
| $(10^{-4}, 1, 10^{-4})$ | $19.4 \pm 0.4$ |
| $(10^{-2}, 1, 10^{-4})$ | $19.2 \pm 0.4$ |
| $(1, 1, 10^{-4})$ | $\mathbf{1.90 \pm 0.27}$ |

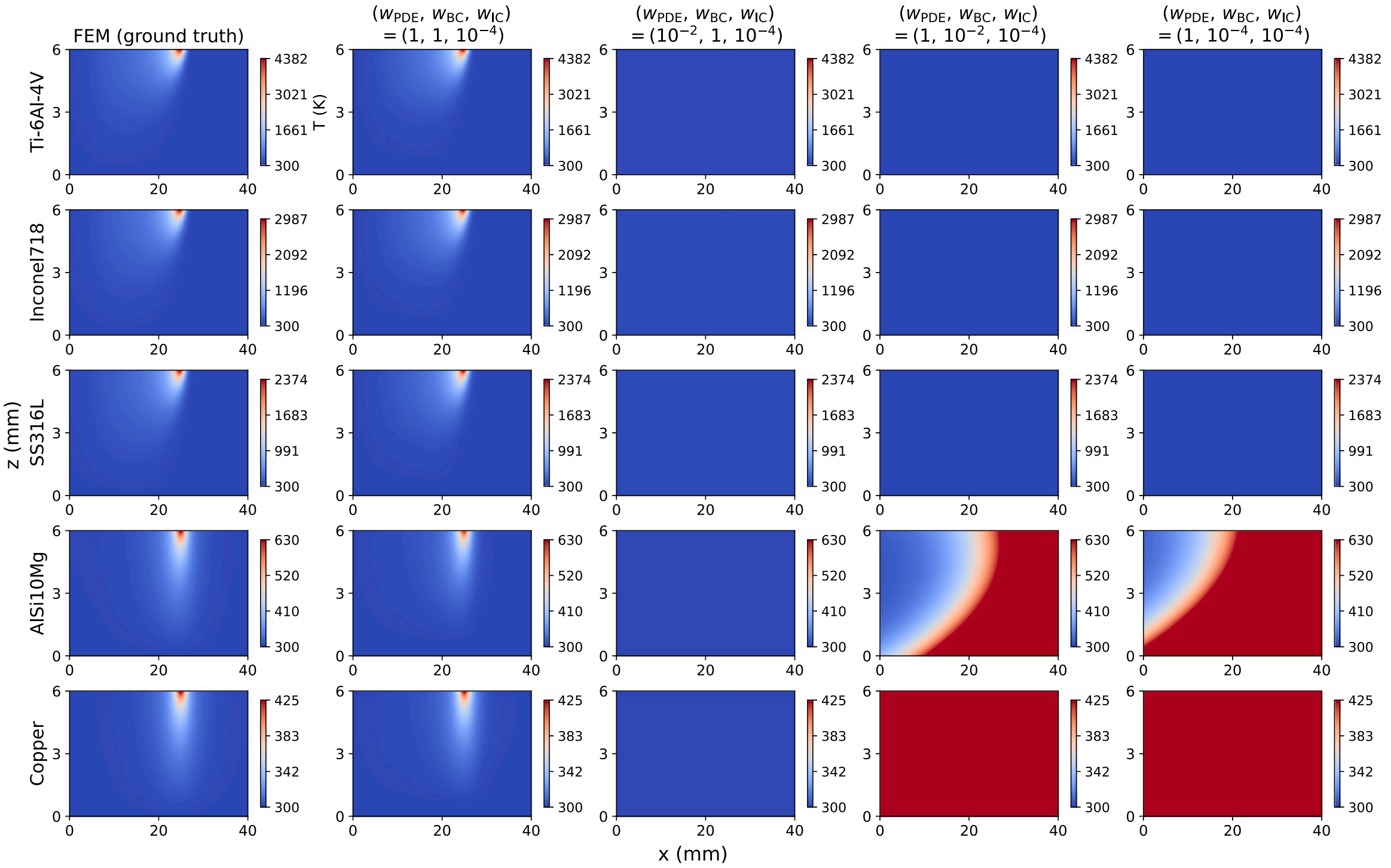

**Fig. A.19.** Cross section temperature field at $t$ = 2.0 s for the five evaluation materials (rows). Columns: FEM ground truth and PINN predictions for different loss weights $(w_{\text{pde}}, w_{\text{bc}}, w_{\text{ic}})$. All loss weight combinations other than the baseline $(w_{\text{pde}}, w_{\text{bc}}, w_{\text{ic}}) = (1, 1, 10^{-4})$ collapses the solution.

**Table 13**
Ablation study on the Adam/L-BFGS epoch split: mean relative $L_2$ error (%) over 5 seeds versus the number of Adam epochs (L-BFGS receives the remaining 10,000− Adam epochs). The 2000-epoch configuration achieves the lowest error.

| Adam epochs | 0 | 500 | 1000 | 2000 | 4000 | 8000 |
|---|---|---|---|---|---|---|
| $L_2$ Error (%) | $3.29 \pm 0.42$ | $3.36 \pm 1.08$ | $3.92 \pm 0.62$ | $\mathbf{1.90 \pm 0.27}$ | $3.56 \pm 1.54$ | $3.10 \pm 1.07$ |

Crucially, we observed that PINN models tend to visit this trivial state first before navigating toward the physically correct solution. For instance, the proposed framework followed the second trajectory during the first 2000 epochs, where the Adam optimizer was applied, and then converged toward the first trajectory once our hybrid optimization strategy was applied. In N-PINN, the model followed the second trajectory until around 18,000 epochs, and then converged toward the first trajectory.

Based on these empirical insights, rather than introducing additional computational overhead via dynamic loss balancing algorithms (e.g., adaptive loss balancing), we deliberately adopted the fixed weight configuration of $(w_{\text{pde}}, w_{\text{bc}}, w_{\text{ic}}) = (1, 1, 10^{-4})$, which aligns with the intrinsic scaling of a well-behaved optimization path.

To further validate this claim, we conducted comprehensive sensitivity analyses across nine different loss weight combinations, varying PDE loss and BC loss weights among $10^{-4}$, $10^{-2}$, and 1. Across five independent random seeds, only the configuration $(w_{\text{pde}}, w_{\text{bc}}, w_{\text{ic}}) = (1, 1, 10^{-4})$ consistently converged to a physically consistent, non-trivial solution (see Table 12).

As shown in Figs. A.18 and A.19, qualitative results further confirmed that while our selected configuration successfully captures the complex thermal behaviors, all other combinations failed by collapsing into poor local optima, such as predicting a uniform ambient temperature field.

### *A.6. Analysis of training epochs in hybrid optimization strategy*

In the proposed hybrid optimization strategy, the number of Adam training epochs prior to transitioning to L-BFGS is indeed a key hyperparameter. Through empirical evaluations, we determined this transition point to be 2000 epochs (see Table 13).

## Data availability

The data and code are available at https://github.com/hsleecri/MaterialAgnosticTempPred.